\documentclass{article}
\usepackage{arxiv}
\usepackage[svgnames]{xcolor}
\usepackage[utf8]{inputenc} 
\usepackage[T1]{fontenc}    
\usepackage[colorlinks=true, citecolor=blue]{hyperref}       
\usepackage{url}            
\usepackage{booktabs}       
\usepackage{amsfonts}       
\usepackage{nicefrac}       
\usepackage{microtype}      
\usepackage{lipsum}		
\usepackage{graphicx}
\usepackage{natbib}
\usepackage{doi}

\input{macros}

\allowdisplaybreaks
\title{The Limits and Potentials of Local SGD for Distributed Heterogeneous Learning with Intermittent Communication}

\author{
	Kumar Kshitij Patel\thanks{Corresponding author. This paper is an extended version of a conference paper at the 37th Annual Conference on Learning Theory (COLT 2024). Preliminary versions of this work also appeared at the Federated Learning and Analytics in Practice: Algorithms, Systems, Applications, and Opportunities workshop at ICML 2023 and the 15th International OPT Workshop on Optimization for Machine Learning at NeurIPS 2023.}\\
	Toyota Technological Institute, Chicago\\
	\texttt{kkpatel@ttic.edu} \\
        \And
	Margalit Glasgow\\
        Stanford University \\
	\texttt{mglasgow@stanford.edu}\\
        \And
	Ali Zindari\\
	CISPA Helmholtz Center\\
	\texttt{ali.zindari@cispa.de}\\
        \And
	Lingxiao Wang\\
	Toyota Technological Institute, Chicago\\
	\texttt{lingxw@ttic.edu}\\
        \And
	Sebastian U. Stich\\
	CISPA Helmholtz Center\\
	\texttt{stich@cispa.de}\\
        \And
	Ziheng Cheng\\
        Peking University\\
	\texttt{alex-czh@stu.pku.edu.cn}\\
        \And
	Nirmit Joshi\\
	Toyota Technological Institute, Chicago\\
	\texttt{nirmit@ttic.edu}\\
        \And
	Nathan Srebro\\
	Toyota Technological Institute, Chicago\\
	\texttt{nati@ttic.edu}\\
}

\date{}

\renewcommand{\headeright}{}
\renewcommand{\undertitle}{}
\renewcommand{\shorttitle}{Local SGD for Heterogeneous Distributed Learning}

\hypersetup{
pdftitle={Local SGD for Distributed Heterogeneous Learning},
pdfsubject={q-bio.NC, q-bio.QM},
pdfauthor={Kumar Kshitij Patel},
pdfkeywords={Distributed Optimization, Local SGD, Intermittent Communication, Min-max Optimal},
}

\begin{document}
\maketitle

\begin{abstract}
    Local SGD is a popular optimization method in distributed learning, often outperforming other algorithms in practice, including mini-batch SGD. Despite this success, theoretically proving the dominance of local SGD in settings with reasonable data heterogeneity has been difficult, creating a significant gap between theory and practice \citep{wang2022unreasonable}. In this paper, we provide new lower bounds for local SGD under existing first-order data heterogeneity assumptions, showing that these assumptions are insufficient to prove the effectiveness of local update steps. Furthermore, under these same assumptions, we demonstrate the min-max optimality of accelerated mini-batch SGD, which fully resolves our understanding of distributed optimization for several problem classes. Our results emphasize the need for better models of data heterogeneity to understand the effectiveness of local SGD in practice. Towards this end, we consider higher-order smoothness and heterogeneity assumptions, providing new upper bounds that imply the dominance of local SGD over mini-batch SGD when data heterogeneity is low.
\end{abstract}


\section{Introduction}\label{sec:intro}
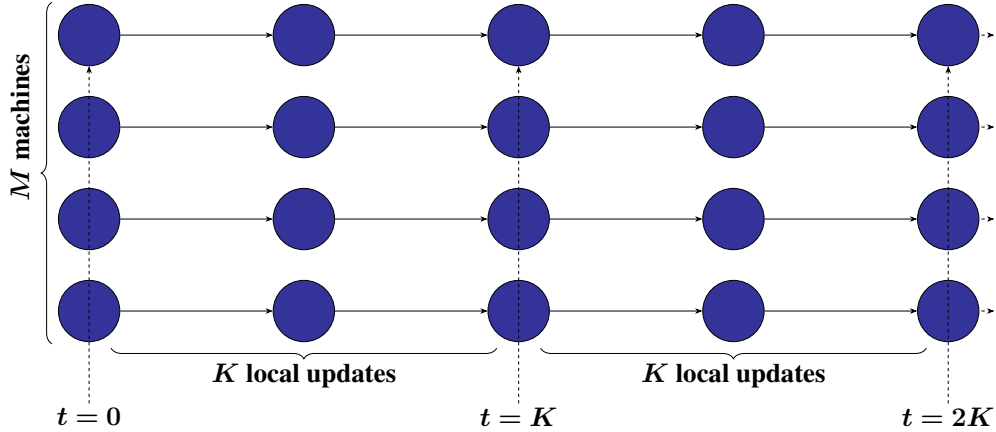
\begin{figure}
    \centering
    \resizebox{0.8\linewidth}{!}{

\begin{tikzpicture}[
    node distance=2cm and 3.5cm,
    line/.style={-Stealth, thick},
    node/.style={circle, draw, thick, fill=NavyBlue!80!White, minimum size=2cm},
    dashedline/.style={dashed, thick},
    timeline/.style={dashed, thick, -Stealth},
    continueline/.style={dashed, thick, -Stealth, shorten <=2pt},
    brace/.style={decorate, decoration={brace, amplitude=10pt, raise=4pt}, thick}
]

\foreach \x in {1,...,5} {
    \foreach \y in {1,...,4} {
        \node[node] (N-\x-\y) at (\x*7, -\y*3) {};
    }
}

\foreach \x [remember=\x as \lastx (initially 1)] in {2,...,5} {
    \foreach \y in {1,...,4} {
        \draw[line] (N-\lastx-\y) -- (N-\x-\y);
    }
}


\draw[brace] ([yshift=-10pt, xshift=-10pt]N-1-4.south west) -- ([yshift=10pt, xshift=-10pt]N-1-1.north west);
\node[left=0.8cm of N-1-2, anchor=south, rotate=90, align=center] {\Huge $\bm{M}$ \textbf{machines}}; 



\node[below=of N-1-4] (t0) {\Huge $\bm{t=0}$};
\draw[timeline] (t0) -- (N-1-1);

\draw[brace, decoration={mirror}] ([yshift=-10pt]N-1-4.south east) -- ([yshift=-10pt]N-3-4.south west) node[midway, below=15pt, align=center] {\Huge$\bm{K}$ \textbf{local updates}};

\node[below=of N-3-4] (t2) {\Huge $\bm{t=K}$};
\draw[timeline] (t2) -- (N-3-1);

\draw[brace, decoration={mirror}] ([yshift=-10pt]N-3-4.south east) -- ([yshift=-10pt]N-5-4.south west) node[midway, below=15pt, align=center] {\Huge$\bm{K}$ \textbf{local updates}};

\node[below=of N-5-4] (t4) {\Huge $\bm{t=2K}$};
\draw[timeline] (t4) -- (N-5-1);

\foreach \y in {1,...,4} {
    \draw[continueline] (N-5-\y) -- ++(1.5, 0);
}


\end{tikzpicture}
}
    \caption{Illustration of the intermittent communication setting.}
    \label{fig:IC}
\end{figure}

We consider the following distributed optimization problem on $M$ machines,
\begin{align}\label{prob:scalar}
    \min_{x\in \rr^d}\rb{F(x):=\frac{1}{M}\sum_{m\in[M]}F_m(x)},
\end{align}
where $F_m := \ee_{z_m \sim \ddd_m}[f(x;z_m)]$ is a stochastic objective on machine $m$, defined using a smooth, convex and differentiable loss function $f(\cdot;z\in \zzz)$ and a data distribution $\ddd_m\in \Delta(\zzz)$. Problem \eqref{prob:scalar} is ubiquitous in machine learning---from training in a data center on multiple GPUs \citep{krizhevsky2012imagenet}, to decentralized training on millions of devices \citep{mcmahan2016federated, mcmahan2016communication}. Perhaps the simplest, most basic, and most important distributed setting for solving Problem \eqref{prob:scalar} is that of intermittent communication (IC) \citep{woodworth2018graph}, where $M$ machines work in parallel over $R$ communication rounds to optimize objective \eqref{prob:scalar}, and during each round of communication, each machine may sequentially compute $K$ stochastic gradient estimates (see Figure \ref{fig:IC} for an illustration). 

While several stochastic gradient descent (SGD) methods have been proposed for solving problem \eqref{prob:scalar} in the IC setting, most of them \citep{kairouz2019advances, wang2021field} are variants of \textbf{local SGD} (update \eqref{eq:local_updates}) or \textbf{mini-batch SGD} (update \eqref{eq:MB_SGD}). In local SGD, each machine computes $K$ sequential stochastic gradients in parallel, and then the resulting updates are averaged across the machines at each round of communication. In practice, this simple algorithm often outperforms all other first-order optimization algorithms \citep{charles2021large,wang2022unreasonable}, including mini-batch SGD \citep{lin2018don, woodworth2020local}. As such, there has been a concerted effort to understand the effectiveness of local SGD for more than a decade \citep{mcdonald2009efficient, zinkevich2010parallelized, zhang2016parallel, stich2018local, dieuleveut2019communication, khaled2020tighter, koloskova2020unified, woodworth2020local, karimireddy2020scaffold, woodworth2020minibatch, yuan2020federated, woodworth2021min, glasgow2022sharp, wang2022unreasonable, patel2023federated}.

In the homogeneous setting, when $\ddd_m=\ddd$ for each $m\in[M]$, \citet{woodworth2021min} showed that the min-max optimal algorithm for optimizing problem \eqref{prob:scalar} with smooth and convex objectives, is the best of (accelerated) local and mini-batch SGD. Unfortunately, their results also implied that local SGD can outperform mini-batch SGD only when even SGD on a single machine (i.e., without any collaboration) also outperforms mini-batch SGD! This result was surprising because, in practice, local SGD almost always outperforms both single-machine SGD and mini-batch SGD \citep{mcmahan_ramage_2017, charles2021large}. The most likely reason for this disparity between theory and practice is that the homogeneous setting is too simplistic. In practice, the data distributions of different machines are ``similar" but not the same, making collaboration pivotal. 

Towards this end, several works \citep{khaled2020tighter, karimireddy2020scaffold, koloskova2020unified, woodworth2020minibatch, yuan2020federated, glasgow2022sharp, wang2022unreasonable} have tried to capture the similarity between machine's distributions by using  \textit{``data heterogeneity''} assumptions. However, most of these works either fail to show a theoretical advantage of local SGD over mini-batch SGD \citep{khaled2020tighter, karimireddy2020scaffold, koloskova2020unified} or they need to make restrictive assumptions that do not allow any interesting data heterogeneity (c.f., Assumption \ref{ass:zeta_everywhere} of \citet{woodworth2020minibatch}). This inability to theoretically explain the effectiveness of local SGD presents a big gap between the theory and practice of distributed optimization. Recently, \citet{wang2022unreasonable} called this the \textbf{\textit{``unreasonable effectiveness of local SGD''}}. In this paper, we contribute to this discourse about the usefulness of local SGD by providing new lower and upper bounds for solving problem \eqref{prob:scalar} in the IC setting. Our key theoretical insights are as follows:
 
\paragraph{1. Existing first-order heterogeneity assumptions are insufficient for local SGD.} We provide a new lower bound for local SGD (Theorem \ref{thm:new_LSGD_lower_bound}) in the general convex setting. Our core finding can be stated as:
\begin{quotation}
    \noindent\textit{\textbf{There is a smooth, convex, and quadratic problem instance such that local SGD can not get arbitrarily close to the shared optimum of the clients with finite communication.}}
\end{quotation}    
Our lower bound precludes dominance of local SGD over mini-batch SGD under the following heterogeneity assumption considered by several existing works \citep{koloskova2020unified, woodworth2020minibatch, glasgow2022sharp}. 
\begin{assumption}[Bounded First-Order Heterogeneity at Optima]\label{ass:zeta_optimal}
    A set of objectives $\{F_m\}_{m\in[M]}$ satisfy $\zeta_\star$-first-order heterogeneity at the optima if for all minimizers of the average objective $x^\star\in \arg\min_{x\in \rr^d} F(x)$, 
    $$\frac{1}{M}\sum_{m\in[M]}\norm{\nabla F_m(x^\star)}^2 \leq \zeta_\star^2.$$
\end{assumption}
We show that the above assumption, while a natural measure of data-heterogeneity, can not avoid some basic pathologies of distributed learning with convex objectives. Our lower bound also precludes any dominance of local SGD over mini-batch SGD under another similar assumption postulated by \citet{wang2022unreasonable} (Assumption \ref{ass:rho} in Appendix \ref{app:wang}).

\paragraph{2. Accelerated mini-batch SGD is min-max optimal when machines have shared optima.} We provide a new algorithm-independent lower bound (Theorem \ref{thm:AIlb_zeta0}) under Assumption \ref{ass:zeta_optimal} (and Assumption \ref{ass:rho}), showing that accelerated mini-batch SGD is min-max optimal under this assumption! This conclusion is surprising, and while it further deepens the theory-practice disparity, it concludes a recent line of work on identifying the min-max optimal algorithm in the general convex setting under related first-order heterogeneity assumptions \citep{khaled2020tighter, koloskova2020unified, woodworth2020minibatch,  glasgow2022sharp, wang2022unreasonable}.

Together our lower bounds (Theorems \ref{thm:new_LSGD_lower_bound} and \ref{thm:AIlb_zeta0}) imply that existing first-order assumptions are insufficient to understand the effectiveness of local SGD, in that they do not characterize a measure of data heterogeneity, reducing which would allow local update steps to be useful. This motivates us to consider higher-order assumptions in addition to first-order ones. 

\paragraph{3. Local SGD shines under higher-order heterogeneity and smoothness assumptions.} We provide a new upper bound (Theorem \ref{col: convex upper bound}) for local SGD improving the analysis of \citet{woodworth2020minibatch} by capturing the effect of second-order heterogeneity (Assumption \ref{ass:tau}) and third-order smoothness (Assumption \ref{ass: lipschitz hessian}). While our upper bound depends on the restrictive Assumption \ref{ass:zeta_everywhere} of \citet{woodworth2021min}, it relaxes local SGD's dependence on the assumption, i.e., our analysis implies a benefit of local updates with much larger data heterogeneity than that implied by the analysis of \citet{woodworth2020minibatch}. This hints that higher-order assumptions indeed enhance the benefit of local updates. We explore this further by considering the special case when all the client objectives are strongly convex quadratics. In this setting, we provide a new upper bound (Theorem \ref{thm:fixed_convergence}) for local SGD, underlining that Assumptions \ref{ass:zeta_optimal} and \ref{ass:tau} control the discrepancy between the fixed point of local SGD and the minimizer of $F$ in problem \eqref{prob:scalar}.


\paragraph{Notation.} We use $\cong$, $\preceq$, and $\succeq$ to refer to equality and inequality up to absolute numerical constants. We denote the set $\{i, i+1, \dots, n\}$ by $[i,n]$ and when $i=1$ by $[n]$. $\norm{.}$ refers to the $L_2$ norm when applied to vectors in $\rr^d$ and the spectral norm when applied to matrices in $\rr^{d\times d}$.

\section{Setting and Preliminaries}\label{sec:setting}
{\renewcommand{\arraystretch}{1.5}%
\begin{table*}
    \centering
    \resizebox{\linewidth}{!}{
    \begin{tabular}{l|l}
    \toprule[2pt]
        \rowcolor[gray]{.8} Type of Result (Reference) & Convergence bound on $\ee[F(\hat x) - F(x^\star)] \preceq/\succeq$ \\
        \midrule[2pt]
        Local SGD Upper Bound & \multirow{2}{*}{$\boldsymbol{\frac{HB^2}{R}} + \frac{(H\sigma^2B^4)^{1/3}}{K^{1/3}R^{2/3}} 
+ \frac{\sigma B}{\sqrt{MKR}} +  \frac{(H\zeta_\star^2B^4)^{1/3}}{R^{2/3}}$}\\
        \citep{koloskova2020unified} & \\
        \hline
        Local SGD Lower Bound & \multirow{2}{*}{$\frac{HB^2}{KR} + \min\cb{\frac{\sigma B}{\sqrt{KR}}, \frac{(H\sigma^2B^4)^{1/3}}{K^{1/3}R^{2/3}}} 
+ \frac{\sigma B}{\sqrt{MKR}} + \min \cb{\frac{\zeta_\star^2}{H}, \frac{(H\zeta_\star^2B^4)^{1/3}}{R^{2/3}}}$}\\
        \citep{glasgow2022sharp} & \\
        \hline
        Local SGD Lower Bound & \multirow{2}{*}{$\boldsymbol{\frac{HB^2}{R}} + \frac{(H\sigma^2B^4)^{1/3}}{K^{1/3}R^{2/3}} 
+ \frac{\sigma B}{\sqrt{MKR}} + \frac{(H\zeta_\star^2B^4)^{1/3}}{R^{2/3}}$}\\
        (Theorem \ref{thm:new_LSGD_lower_bound}) & \\
        \midrule[2pt]
        Mini batch SGD Upper Bound & \multirow{2}{*}{$\boldsymbol{\frac{HB^2}{R}}
+ \frac{\sigma B}{\sqrt{MKR}}$}\\
        \citep{dekel2012optimal} &\\
        \hline
        Acc. MB SGD Upper Bound & \multirow{2}{*}{$\boldsymbol{\frac{HB^2}{R^2}}
+ \frac{\sigma B}{\sqrt{MKR}}$}\\
        \citep{ghadimi2012optimal} &\\
        \hline
        Algo. Independent Lower Bound & \multirow{2}{*}{$\min \cb{\boldsymbol{\frac{HB^2}{R^2}}, \frac{\zeta_\star^2}{HR^2}, \frac{\zeta_\star B}{R}} 
+ \frac{\sigma B}{\sqrt{MKR}}$}\\
        \citep{woodworth2020minibatch} & \\
        \hline
        Algo. Independent Lower Bound & \multirow{2}{*}{$\boldsymbol{\frac{HB^2}{R^2}} + \frac{\sigma B}{\sqrt{MKR}}$}\\
        (Theorem \ref{thm:AIlb_zeta0}) & \\
        \bottomrule[2pt]
    \end{tabular}
    }
    \caption{Summary of old and new results (up to logarithmic factors) for problem instances in the class $\ppp_{\zeta_\star}^{H,B,\sigma}$, i.e., under Assumption \ref{ass:zeta_everywhere} on $S^\star$. Throughout we assume $\frac{\sigma B}{\sqrt{MKR}} \leq HB^2$.}
    \label{tab:summary}
\end{table*}}

Note that an instance of problem \eqref{prob:scalar} can be characterized by the client distributions $\{\ddd_m\in \Delta(\zzz)\}_{m\in[M]}$ and a differentible loss function $f(\cdot;z\in\zzz):\rr^d\to \rr$. We will denote the set of all such problem instances by $\ppp$. To further restrict the problems we will study, we assume that for all $m\in[M]$, the objective function $F_m$ is convex, and $H$-smooth, i.e., for all $m\in[M],\ x,y\in\rr^d$, 
\begin{align}\label{ass:smth}
    F_m(x) \leq F_m(y) + \inner{\nabla F_m(y)}{x-y} + \frac{H}{2}\norm{x-y}^2\enspace.
\end{align}
We also assume that each machine $m\in[M]$ computes stochastic gradients of $F_m$ in the IC setting by sampling from its distributions $z\sim\ddd_m$, and that for all $x\in\rr^d$,
\begin{align}\label{ass:stoch_first_order}
    \ee_{z\sim \ddd_m}\sb{\nabla f(x;z)}=\nabla F_m(x)\enspace, \quad\text{ and }\quad \ee_{z\sim \ddd_m}\sb{\norm{\nabla f(x; z) - \nabla F_m(x)}^2} \leq \sigma^2\enspace.
\end{align}
Finally, we assume that the average objective $F$ has bounded optima, i.e., 
\begin{align}\label{ass:bounded_optima}
    \norm{x^\star}\leq B,\ \forall\ x^\star\in S^\star := \arg\min_{x\in \rr^d} F(x)\enspace.
\end{align}
We will denote the class of all the problems satisfying these assumptions\footnote{Note that when $\sigma=0$, i.e., in the noiseless setting, then the problem instance is characterized by $\{F_m\}_{m\in[M]}$.} by $\ppp^{H, B, \sigma}$. With this, we are ready to describe our algorithms with intermittent communication. We can write the update for local SGD (initialized at $x_0$) for round $r\in[R]$ as follows,
\begin{equation}\label{eq:local_updates}
    \begin{aligned}
    x_{r,0}^m &= x_{r-1}, && \forall\ m\in[M]\\
    x_{r,k+1}^m &= x_{r,k}^m - \eta \nabla f(x_{r,k}; z_{r,k}^m),\ z_{r,k}^m\sim \ddd_m, && \forall\ m\in[M], k\in[0, K-1]\\
    x_{r} &= x_{r-1} + \frac{\beta}{M}\sum_{m\in[M]}\rb{x_{r,K}^m-x_{r-1}}.   
    \end{aligned}
\end{equation}
Above $x_{r,k}^m$ is the $k^{th}$ local model on machine $m$, leading up to the $r^{th}$ round of communication, while $x_r$ is the consensus model at the end of the $r^{th}$ communication. For local SGD, $\eta$ is referred to as the inner step size, while $\beta$ is the outer step size. Setting $\beta=1$ recovers \textbf{\textit{``vanilla local SGD''}} with a single step size which has been analyzed in several earlier works \citep{stich2018local, dieuleveut2019communication, khaled2020tighter, woodworth2020local}. Vanilla local SGD is equivalent to averaging the machine's models after $K$ local updates. Similarly, we can write the updates of mini-batch SGD (initialized at $x_0$) for round $r\in[R]$ as follows,
\begin{equation}\label{eq:MB_SGD}
    \begin{aligned}
    g_{r,k}^m &= \nabla f(x_{r-1}; z_{r,k}^m),\ z_{r,k}^m\sim \ddd_m, && \forall\ m\in[M], k\in[0, K-1]\\
    x_{r} &= x_{r-1} - \frac{\beta}{M}\sum_{m\in[M], k\in[0, K-1]}g_{r,k}^m.
    \end{aligned}
\end{equation}
 The main difference in the mini-batch update compared to vanilla local SGD is that its local gradient is computed at the same point for the entire communication round. Due to this, mini-batch SGD is not impacted by data heterogeneity, as it optimizes $F$ without getting affected by the multi-task nature of problem \eqref{prob:scalar} (for more discussion, see \cite{woodworth2020minibatch}). However, this is also why local SGD can intuitively outperform mini-batch SGD: it has more effective updates than mini-batch SGD. For e.g., without noise, i.e., $\sigma=0$, mini-batch SGD keeps obtaining the gradient at the same point, thus making only $R$ updates instead of $KR$ updates of local SGD. 

When the data heterogeneity is ``low'',  the iterates on different machines would not go too far between communication rounds, a.k.a., local SGD will have a small ``consensus error''. In the extreme homogeneous setting, the iterates on different machines will be the same in expectation. We will denote the class of homogeneous problems by $\ppp_{hom}^{H, B, \sigma}\subset \ppp^{H, B, \sigma}$. Most heterogeneity assumptions are variants of first-order assumptions, which impose restrictions on the gradients of functions and thus aid the analysis of local SGD by controlling how far the machines' iterates can go between two communication rounds. These assumptions can be generalized as follows.

\begin{assumption}[Bounded First-Order Heterogeneity]\label{ass:zeta_everywhere}
    A set of objectives $\{F_m\}_{m\in[M]}$ satisfy $\zeta$-first-order heterogeneity on the domain $\xxx\subseteq\rr^d$ if, 
    $$\sup_{x\in \xxx, m\in[M]}\norm{\nabla F_m(x) - \nabla F(x)}^2 \leq \zeta(\xxx)^2.$$
\end{assumption}
We will refer to problems satisfying Assumption \ref{ass:zeta_everywhere} on $\xxx$ by $\ppp_{\zeta(\xxx)}^{H,B, \sigma}$ and omit the dependence on $\xxx$ when it is clear from the context. \citet{woodworth2020minibatch} showed an advantage of vanilla local SGD over mini-batch SGD under Assumption \ref{ass:zeta_everywhere} on the unbounded domain $\xxx=\rr^d$ (see Table \ref{tab:summary2}). Unfortunately, assuming $\xxx$ to be $\rr^d$ makes the above assumption too restrictive, for e.g., quadratic functions in the class $\ppp^{H,B,\sigma}_{\zeta(\rr^d)}$ must have the same Hessian. 
\begin{proposition}\label{prop:zeta_limitation}
Let $F_m(x)=\frac{1}{2}x^TA_mx + b_m^Tx + c_m$ for all $m\in[M]$. If $\{F_m\}_{m\in[M]}$ satisfy Assumption \ref{ass:zeta_everywhere} for $\xxx=\rr^d$ for $\zeta<\infty$ then for any two machines $m, n\in[M]$, $A_m=A_n$.   
\end{proposition}
\begin{remark}[$\bm{\ppp_{hom}^{H,B, \sigma=0}\approx \ppp_{\zeta(\rr^d)<\infty}^{H,B, \sigma=0}}$]
    Essentially, Assumption \ref{ass:zeta_everywhere} on $\rr^d$ only allows heterogeneity in linear terms. In the noiseless setting, i.e., when $\sigma=0$, such heterogeneity can be adjusted for using one round of communication at the beginning of optimization by computing the gradient at $0$ on all machines and transmitting $\nabla F(0)$ to the machines. At each round, the machines add the correction $\nabla F(0)-\nabla F_m(0)$ to their gradients and behave as if the problem were homogeneous!
\end{remark}
The restrictiveness of Assumption \ref{ass:zeta_everywhere} with $\xxx=\rr^d$ is precisely why several works \citep{khaled2020tighter, karimireddy2020scaffold, koloskova2020unified} have considered relaxed versions of the assumption. A natural relaxation is to assume Assumption \ref{ass:zeta_everywhere} on $\xxx=S^\star$, the set of optima of $F$, which is precisely Assumption \ref{ass:zeta_optimal}. We will denote $\zeta_\star := \zeta(S^\star)$, and define the problem class of problems satisfying Assumption \ref{ass:zeta_optimal} by $\ppp_{\zeta_\star}^{H,B,\sigma} := \ppp_{\zeta(S^\star)}^{H,B,\sigma}$. We can make several interesting remarks related to interpretations of Assumption \ref{ass:zeta_optimal}.
\begin{remark}[$\bm{\ppp^{H,B,\sigma=0}_{hom}\approx\ppp_{\zeta(\rr^d)<\infty}^{H,B, \sigma=0}\subset \ppp_{\zeta_\star=0}^{H,B,\sigma=0}}$]\label{rem:key_insight}
    Note that $\zeta_\star=0$ does not imply we are in the homogeneous regime---all it says is that the machines should have some shared optimum. This is in contrast to Assumption \ref{ass:zeta_everywhere} with $\xxx=\rr^d$, where $\zeta=0$ implies that $F_m$'s have to be the same except for constants. Thus, assuming $\zeta_\star <\infty$ (or even $\zeta_\star=0$) \textbf{does} allow for interesting data heterogeneity. 
\end{remark}
\begin{remark}[Approximate Simultaneous Realizability]\label{rem:realizability}
    In a learning setting, Assumption \ref{ass:zeta_optimal} can be interpreted as an approximate \emph{simultaneous realizability} assumption for $\ddd_m$'s, measuring how good the solution for the average objective $F$ is for any individual machine. Besides literature on data heterogeneity in distributed optimization, the simultaneous realizability assumption also appears in the collaborative PAC learning literature \citep{blum2017collaborative, nguyen2018improved, haghtalab2022demand} and in the context of avoiding defections in federated learning \citep{han2023effect}.
\end{remark}
\begin{remark}[Local SGD's Fixed Point]\label{rem:fixed}
    Denote the set of optima of machine $m\in[M]$ by $S_m^\star := \arg\min_{x\in\rr^d}F_m(x)$. Assume the machines have some shared optimum $x^\star \in \cap_{m\in[M]}S_m^\star$, then in fact all their optima should be shared, i.e., $S^\star = \cap_{m\in[M]}S_m^\star$. This observation implies that all points in $S^\star = \cap_{m\in[M]}S_m^\star$ are fixed points of local gradient descent with any $K\geq 1$ \citep{pathak2020fedsplit}. Interestingly beyond the setting of shared optima, for $K>1$ there is no simple closed form for the fixed points of local gradient descent. Since $\zeta_\star=0$ implies that the machines do have a shared optimum, Assumption \ref{ass:zeta_optimal} intimately ``controls'' the convergence behavior of local gradient descent.   
\end{remark}

Hopefully, the above remarks motivate the importance of understanding local SGD's performance on the problem class $\ppp_{\zeta_\star}^{H,B, \sigma}$. Unfortunately, no known analysis has provably shown that local SGD improves over mini-batch SGD on this class (see Table \ref{tab:summary}). Based on the best known lower bound by \citet{glasgow2022sharp} (see Table \ref{tab:summary}), it is still \em possible \em that local SGD could improve upon (accelerated) mini-batch SGD when $\zeta_\star$ and $\sigma$ are small, but $K$ is large. For instance, in the extreme case when $\sigma=\zeta_\star=0$, even if $K\to\infty$ but $R$ is small, accelerated large mini-batch SGD will not have zero function sub-optimality. In contrast, if the lower bound by \citet{glasgow2022sharp} were tight, local SGD would get zero function sub-optimality in such a regime. This raises the following question: 
\begin{quote}
    \textbf{\textit{Can local SGD dominate mini-batch SGD on the class $\ppp_{\zeta_\star}^{H,B, \sigma}$ for small $\zeta_\star$?}}    
\end{quote}
 We will answer this question negatively in the next section, and also establish the min-max optimality of mini-batch SGD amongst the class of distributed zero-respecting algorithms in the IC setting on the problem class $\ppp_{\zeta_\star}^{H,B, \sigma}$.

\section{Middling Utility of First-order Heterogeneity Assumptions in the Convex Setting}\label{sec:lower}
Our lower bounds in this section are based on the key insight presented in Remark \ref{rem:key_insight}, i.e., even if the machines have a shared optimum, i.e., $S^\star = \cap_{m\in[M]}S_m^\star\neq \emptyset$ (see Remark \ref{rem:fixed}), the problem instance need not be homogeneous. As a motivating example, assume there are two machines, we do not have any noise, i.e., $\sigma=0$ and the objectives of the machines for all $x=(x[1],x[2])\in \rr^2$ are given by,
\begin{equation} \label{eq:motivation}
\begin{aligned}
    F_1(x) = \frac{H}{2}\rb{x[1]-x^\star[1]}^2 \text{ and } F_2(x) = \frac{H}{2}\rb{x[2]-x^\star[2]}^2,
\end{aligned}    
\end{equation}
where $x^\star=(x^\star[1],x^\star[2]) \in \rr^2$ is the unique optimum of $F(x) = \frac{H}{4}\norm{x-x^\star}^2$. Note that each machine's objective is $H$-smooth and satisfies Assumption \ref{ass:zeta_everywhere} on $S^\star=\{x^\star\}$ with $\zeta_\star=0$. Assuming we run local SGD on both machines initialized at $(0,0)$, then the iterate after $R$ rounds is given by (proved in Appendix \ref{app:lower}), 
\begin{align}\label{eq:lsgd_motivation}
  x_R = x^\star\rb{1 - \rb{1- \frac{\beta}{2}\rb{1 - (1-\eta H)^K}}^R}.  
\end{align}
Let us first consider vanilla local SGD, i.e., $\beta=1$. The above expression simplifies to,
\begin{align}
  x_R = x^\star\rb{1 - \rb{\frac{1 + (1-\eta H)^K}{2}}^R}.  
\end{align}
Thus, even if $K\to\infty$, $x_R$ \textbf{does not} converge to $x^\star$ for finite $R$. This implies that the lower bound from \citet{glasgow2022sharp}, which goes to $0$ when $K\to\infty$, must be loose.
\begin{remark}[The Role of the Outer Step-size]
        Note that for Example \ref{eq:motivation}, if we set $\beta=2$, $x_R$ will converge to $x^\star$ in a single communication round when $K\to\infty$! Thus, at least in this example, we can not rule out the lower bound by \citet{glasgow2022sharp} if we use the correct outer step size. This example re-emphasizes the role of outer step-size \citep{charles2020outsized, jhunjhunwala2023fedexp} and provides a separation between vanilla local SGD and local SGD with two step-sizes.  
\end{remark}
In the following proposition, we extend this idea to obtain a lower bound for any outer step size $\beta$ (and not just $1$).  
\begin{proposition}\label{prop:lb_zeta0}
For any $K \geq 2, R, M, H, B$, and $\sigma=0$ there exist $\{F_m\}_{m\in[M]}$ in the problem class $\ppp^{H,B, \sigma=0}_{\zeta_\star=0}$, s.t. the local GD iterate $x_R$ with any step-size $\eta, \beta>0$, and initialized at zero has:
\begin{align*}
 \mathbb{E}\left[F(\hat{x}_R)\right] - F(\x^{\star}) &\succeq \frac{HB^2}{R} \,.
\end{align*}
\end{proposition}
The above proposition formalizes the idea presented in our motivating example \eqref{eq:motivation}, showing that local GD can not reach arbitrary function sub-optimality with finite $R$ but infinite $K$ for any $\beta$. In fact, we also use a quadratic construction to prove the proposition. This lower bound also precludes any domination over gradient descent on $F$ with $R$ updates, which is indeed the noiseless variant of mini-batch SGD (see update \eqref{eq:MB_SGD}). Finally, note that the proposition can not benefit from simultaneous realizability, i.e., non-empty $\cap_{m\in[M]}S_m^\star$. This highlights the hardness of general convex optimization in the intermittent communication model. To prove the above proposition for arbitrary step sizes, we reduce the local GD updates over a single communication round to a single gradient descent (GD) update on $F$ and then use the following lemma (proved in Appendix \ref{app:lower}) for GD.
\begin{lemma}\label{lem:condition_number}
    Let $F(x)$ be a convex quadratic function whose Hessian has top eigenvalue $H$, bottom eigenvalue $\mu$, and condition number $\kappa = \frac{H}{\mu} \geq 6$. Let $\hat{x}_R$ be the $R^{th}$ gradient descent iterate initialized at zero. Then for any $B$, there exists some $x^\star$ with $\|x^\star\|_2 \leq B$ such that for any step size $\eta$, $F(\hat{x}_R) - F(x^\star) \geq HB^2 \frac{1}{\kappa} e^{-R/\kappa}$. In particular, if $\kappa = \Omega(R)$, then $F(\hat{x}_R) - F(x^\star) \geq \Omega\left(\frac{HB^2}{R}\right)$.
\end{lemma}
 To get a lower bound in terms of $\zeta_\star$ from Assumption \ref{ass:zeta_everywhere} as well as understand the effect of noise level $\sigma$, we combine proposition \ref{prop:lb_zeta0} with the previous lower bound by \citet{glasgow2022sharp}, to get the following lower bound result. 
\begin{theorem}\label{thm:new_LSGD_lower_bound}
For any $K \geq 2, R, M, H, B, \sigma, \zeta_\star$, there exists a problem instance in $\ppp^{H,B,\sigma}_{\zeta_\star}$ such that the final iterate $x_R$ of local SGD initialized at zero with any step size satisfies:
\begin{align*}
\ee\sb{F(x_R)} - F(x^\star) &\succeq \frac{HB^2}{R} + \frac{(H\sigma^2B^4)^{1/3}}{K^{1/3}R^{2/3}} + \frac{\sigma B}{\sqrt{MKR}} + \frac{(H\zeta_\star^2B^4)^{1/3}}{R^{2/3}}.
\end{align*}
\end{theorem}
Combining Theorem \ref{thm:new_LSGD_lower_bound} with the upper bound by \citet{koloskova2020unified}, we characterize the optimal convergence rate for local SGD for problems in class $\ppp_{\zeta_\star}^{H,B,\sigma}$. Thus, our lower bound concludes the line of work trying to analyze local SGD under the first-order heterogeneity assumption on the set $S^\star$ \citep{koloskova2020unified, woodworth2020minibatch, glasgow2022sharp}. 
\begin{remark}[Proposed Assumption of \citet{wang2022unreasonable}]
    Proposition \ref{prop:lb_zeta0} also implies that the first-order assumption of \citet{wang2022unreasonable} (see Assumption \ref{ass:rho} and discussion in Appendix \ref{app:wang}) can not be used to show a domination of local SGD over mini-batch SGD. This is because when $S^\star\neq \emptyset$, the data heterogeneity assumption of \citet{wang2022unreasonable} implies zero heterogeneity.
\end{remark}
 Overall, the results of this section imply that none of the existing assumptions besides the Assumption \ref{ass:zeta_everywhere} on the unbounded domain $\rr^d$ provides the necessary control on data heterogeneity, for local SGD to beat mini-batch SGD in the general convex setting.

\subsection{The Min-max Optimality of Mini-batch SGD}\label{sec:AILB}
While we can not show the effectiveness of local SGD on the class $\ppp_{\zeta_\star}^{H,B,\sigma}$, can we say which algorithm is min-max optimal for this class of problems? It turns out that the answer is surprisingly simple: accelerated large mini-batch SGD \citep{ghadimi2012optimal}! To show this, we prove the following new algorithm-independent lower bound, which does not improve with a small $\zeta_\star$. 
\begin{theorem}[Algorithm independent lower bound]\label{thm:AIlb_zeta0}
For any $K \geq 2, R, M, H, B, \sigma, \zeta_\star$, there exists a problem instance in the class $\ppp^{H, B, \sigma}_{\zeta_\star}$, s.t. the final iterate $\hat{x}$ of any distributed zero-respecting algorithm (see Appendix \ref{app:zero_respecting}) initialized at zero with $R$ rounds of communication and $K$ stochastic gradient computations per machine per round satisfies,
\begin{align}
 \mathbb{E}\left[F(\hat{x})\right] - F(x^{\star}) \succeq \frac{HB^2}{R^2} + \frac{\sigma B}{\sqrt{MKR}} \,.
\end{align}
\end{theorem}

The above lower bound fully characterizes the min-max complexity of distributed optimization on the class $\ppp_{\zeta_\star}^{H,B,\sigma}$, showing that accelerated mini-batch SGD is the min-max optimal algorithm. This conclusion is surprising, but it closes a recent line of work investigating the intermittent communication setting under Assumption \ref{ass:zeta_everywhere} \citep{khaled2020tighter, karimireddy2020scaffold, koloskova2020unified, woodworth2020minibatch, glasgow2022sharp, wang2022unreasonable}. The above lower bound also implies the min-max optimality of large mini-batch SGD under Assumption \ref{ass:rho} proposed by \citet{wang2022unreasonable} (see Appendix \ref{app:wang}).

Compared to \citet{woodworth2021min} who showed that for problem class $\ppp_{hom}^{H, B, \sigma}$ (and by implication for $\ppp_{\zeta=0}^{H, B, \sigma}$), either accelerated SGD on a single machine or accelerated local SGD can improve over accelerated mini-batch SGD, we have shown that for the problem class $\ppp_{\zeta_\star=0}^{H, B, \sigma}$ such an improvement is not possible\footnote{Note that on Example \eqref{eq:motivation} also showed that single machine SGD must incur a sub-optimality of order $HB^2$, making it much worse than local SGD, and highlighting that $\ppp^{H,B,\sigma}_{hom}$ is a very special sub-class of $\ppp_{\zeta=0}^{H, B, \sigma}$ with regards to the utility of collaboration.}. Thus, we can not hope to show the benefit of using (accelerated) local SGD on any super-class of $\ppp_{\zeta_\star=0}^{H, B, \sigma}$. In other words, we need additional assumptions on top of Assumption \ref{ass:zeta_everywhere} for $\xxx=S^\star$ to avoid the pathologies of our hard instances. In the remainder of this paper we will explore how additional higher-order assumptions can help improve local SGD's convergence.




\section{Beating Mini-batch SGD with Higher-order Heterogeneity and Smoothness}\label{sec:upper}
{\renewcommand{\arraystretch}{1.5}%
\begin{table*}
    \centering
    \resizebox{\textwidth}{!}{
    \begin{tabular}{l|l}
    \toprule[2pt]
        \rowcolor[gray]{.8} Type of Result (Reference) & Convergence bound on $\ee[F(\hat x) - F(x^\star)] \preceq/\succeq$ \\
        \midrule[2pt]
        \rowcolor[gray]{.7} \multicolumn{2}{c}{$\mu$-Strongly Convex $F$}\\
        \midrule[2pt] 
        Local SGD Upper Bound & \multirow{2}{*}{$\frac{HB^2}{HKR + \mu K^2R^2} + \frac{\sigma^2}{\mu MKR} + \frac{H\zeta^2}{\mu^2R^2} + \frac{H\sigma^2}{\mu^2KR^2}$}\\
        \citep{woodworth2020minibatch} & \\
        \hline
        Local SGD Upper Bound & \multirow{2}{*}{$\exp \left(-\frac{\mu KR}{H}\right)HB^2 + \frac{\sigma^2}{\mu M KR} + \frac{Q^2 \sigma^4}{\mu^5 K^2 R^4}$}\\
        \citep{yuan2020federated} \quad($\star$)& \\
        \hline
        Local SGD Upper Bound & \multirow{2}{*}{$ \exp \left(-\frac{\mu KR}{H}\right)HB^2 + \frac{\sigma^2}{\mu M KR}+\frac{Q^2\sigma^4}{\mu^5 K^2R^4} + \frac{Q^2\zeta^4}{\mu^5 R^4}+  \frac{\tau^2\sigma^2}{\mu^3 KR^2}+  \frac{\tau^2\zeta^2}{\mu^3 R^2}$}\\
        (Theorem \ref{thm: LSGD_upper_bound}, Appendix \ref{app:upper proof}) & \\
        \midrule[2pt]
        \rowcolor[gray]{.7} \multicolumn{2}{c}{General Convex $F$}\\
        \midrule[2pt]
        Local SGD Upper Bound & \multirow{2}{*}{$\frac{HB^2}{KR} + \frac{\sigma B}{\sqrt{MKR}} + \frac{(H\zeta^2B^4)^{1/3}}{R^{2/3}} + \frac{(H\sigma^2B^4)^{1/3}}{K^{1/3}R^{2/3}}$}\\
        \citep{woodworth2020minibatch} &\\
        \hline
        Local SGD Upper Bound & \multirow{2}{*}{$\frac{HB^2}{KR} + \frac{\sigma B}{\sqrt{MKR}} + \frac{(Q\sigma^2B^5)^{1/3}}{K^{1/3}R^{2/3}}$}\\
        \citep{yuan2020federated} \quad($\star$)& \\
         \hline
        Local SGD Upper Bound & \multirow{2}{*}{$\frac{HB^2}{KR} + \frac{\sigma B}{\sqrt{MKR}} + \frac{(\tau \sigma B^3)^{1/2}}{K^{1/4}R^{1/2}} + \frac{(\tau \zeta B^3)^{1/2}}{R^{1/2}} + \frac{(Q\sigma^2B^5)^{1/3}}{K^{1/3}R^{2/3}} + \frac{(Q\zeta^2B^5)^{1/3}}{R^{2/3}}$}\\
        (Theorem \ref{col: convex upper bound}) & \\
        \hline
        Alg. Independent Lower Bound & \multirow{2}{*}{$\frac{HB^2}{K^2R^2} + \min\left\{\frac{\sigma B}{\sqrt{MKR}},HB^2\right\} + \min\left\{\frac{HB^2}{R^2 \log^2 M},\frac{\sqrt{Q\sigma}B^2}{K^{1/4}R^2 \log^{7/4}M},\frac{\sigma B}{\sqrt{KR}}  \right\}$}\\
        \citep{woodworth2021min} \quad($\star$)& \\       
        \bottomrule[2pt]
    \end{tabular}
    }
    \caption{Summary of old and new results for the problem class $\ppp_{\tau}^{H,Q,B,\sigma}~\cap~\ppp_{\zeta(\bb_2(D))}^{H,B,\sigma} $ for $\mu$-strongly-convex and convex average objective $F$. Rows with ($\star$) include results for the homogeneous setting.}
    \label{tab:summary2}
\end{table*}}

In the previous section, we showed that we can not prove the effectiveness of local SGD over mini-batch SGD if we consider the entire class $\ppp^{H,B,\sigma}_{\zeta(S^\star)} = \ppp^{H,B,\sigma}_{\zeta_\star}$. On the other hand, recall that \cite{woodworth2020minibatch} did show such a result over the ``nearly homogeneous" class $\ppp^{H,B,\sigma}_{\zeta(\rr^d)}$ (see Table \ref{tab:summary2}). This prompts us to ask the following question: 
\begin{quote}
\textbf{\textit{Can local SGD dominate mini-batch SGD on the class $\ppp^{H,B,\sigma}_{\zeta(\bb_2(D))}$, where $\bb_2(D)$ is the $L_2$ ball of radius $D$ around $0$ for some sufficiently large $D$, and small $\zeta(\bb_2(D))$? }}    
\end{quote}
Intuitively, if our problem has a low (but meaningful) heterogeneity, and we can ensure that the local SGD iterates never leave the ball $\bb_2(D)$, for some large enough $D$, then we can recover the result of \citet{woodworth2020minibatch}, to show the effectiveness of local SGD on a much larger problem class $\ppp_{\zeta(\bb_2(D))}^{H,B,\sigma}$. To enable this, we would need the following second-order heterogeneity assumption, which controls how far the hessians on each machine can be from the average hessian:
\begin{assumption}[Bounded Second-order Heterogeneity]\label{ass:tau}
    A set of doubly-differentiable objectives $\{F_m\}_{m\in[M]}$ satisfy $\tau$-second-order heterogeneity if for all $x\in \rr^d$,
    $$\sup_{m,n\in[M]}\norm{\nabla^2 F_m(x) - \nabla^2 F_n(x)}^2 \leq \tau^2.$$
\end{assumption}
    \begin{remark}
        Assumption \ref{ass:tau} has recently received attention in the non-convex setting \citep{karimireddy2020mime, murata2021bias, patel2022towards}. \citet{patel2022towards} show that it is possible to show the dominance of local update algorithms, essentially local SGD with variance reduction over mini-batch algorithms when $\tau$ is much smaller than $H$.
    \end{remark}  
    Assumption \ref{ass:tau} can always be satisfied by choosing $\tau\geq 2H$ for smooth functions. We will denote the problems satisfying Assumption \ref{ass:tau} by $\ppp_\tau^{H,B,\sigma}$. Given Assumption \ref{ass:tau}, we note that on the set $\bb_2(D)$, we can bound $\zeta$ in terms of $\tau, \zeta_\star$ and $D$ using the following upper bound (proved in Appendix \ref{app:upper proof}). 
\begin{proposition}\label{prop:zeta_bound}
    Given a problem instance in the class $\ppp_{\zeta_\star}^{H,B,\sigma}~\cap~ \ppp_\tau^{H,B,\sigma}$ with objectives $\{F_m\}_{m\in[M]}$,
    \begin{align*}
        \sup_{x\in\bb_2(D), m\in[M]}\norm{\nabla F_m(x)-\nabla F(x)}^2 \leq (\zeta_\star + \tau D)^2.
    \end{align*}
\end{proposition}
The above proposition implies that for some level of first-order heterogeneity $\zeta$ when $\tau$ is small, we can allow for a much larger $D$. Alternatively, if we know our algorithm will be inside a ball $\bb_2(D)$, the smaller the second-order heterogeneity of our problem, the smaller the bound on its first-order heterogeneity. In the extreme case of $\tau=0$, we can replace $\zeta$ with $\zeta_\star$ (i.e., the converse of Proposition \ref{prop:zeta_limitation})! And as a sanity check, in the homogeneous setting the right hand side is zero. Finally, we will also define a third-order smoothness assumption in synergy with Assumption \ref{ass:tau}. 
\begin{assumption}[Third-order Smoothness]\label{ass: lipschitz hessian}
    The average function $F$ has a $Q$ Lipschitz hessian if 
    \begin{equation*}
       \forall\ x,y \in \mathbb{R}^d, \quad \norm{\nabla^2 F(x) - \nabla^2 F(y)} \leq Q \norm{x-y}\enspace.
    \end{equation*}
\end{assumption}
The above assumption intuitively upper bounds the gap between a function and its second-order approximation and is meaningful for smooth functions only when for a given $x,y\in\rr^d$, $H \gggg Q\norm{x-y}$, i.e., the bound implied by smoothness is ``loose''. For a sanity check, note that when $Q=0$, then $F$ must be a quadratic function. To analyze functions with third-order smoothness, we also need control over the fourth moment of the stochastic noise on each machine, akin to existing works \citep{yuan2020federated}. In particular in addition to the assumptions in \eqref{ass:stoch_first_order} we also assume that for each machine $m\in[M]$ and for all $x\in\rr^d$,
\begin{align}\label{ass:stoch_first_order_fourth}
    \ee_{z\sim \ddd_m}\sb{\norm{\nabla f(x; z) - \nabla F_m(x)}^4} \leq \sigma^4\enspace.
\end{align}
We will denote the problems satisfying Assumptions \ref{ass:tau} and \ref{ass: lipschitz hessian} and the above additional fourth moment bound, in addition to other assumptions in the paper by $\ppp^{H,Q,B,\sigma}_\tau$. 
\begin{remark}
     For quadratic problems in the class $\ppp^{H,B, \sigma}_{hom}$, \citet{woodworth2020local} showed that accelerated local SGD is min-max optimal. In particular, their result implies that local SGD can achieve an arbitrary accuracy for a fixed number of communication rounds $R$ by increasing the number of local steps $K$ per round. Given this result, several works \citep{yuan2020federated, bullins2021stochastic} have tried to use higher-order smoothness assumptions to interpolate between the min-max optimal bounds for the quadratic and convex homogeneous setting (Table \ref{tab:summary2}).
\end{remark}
We can obtain the following informal convergence result for the problem class $\ppp^{H,Q,B,\sigma}_\tau\cap \ppp^{H,B,\sigma}_{\zeta_\star}$. We state the full version of the above bound, and the step-size used to obtain it, in Appendix \ref{app:upper proof}.
\begin{theorem}[Informal]\label{col: convex upper bound}
    For any $K,R, M \geq 1$ and $ H, B , Q, \sigma, \tau \geq 0$ consider a problem instance in the class $\ppp^{H,Q,B,\sigma}_\tau~\cap ~\rb{\ppp^{H,B,\sigma}_{\zeta(\bb_2(D))}~\cup~\ppp^{H,B,\sigma}_{\zeta_\star}}$, where we assume $D$ is large-enough such that the local SGD iterates stay inside the ball $\bb_2(D)$. Then, for a fixed inner stepsize of $\eta \leq \frac{1}{2H}$, and outer stepsize of $\beta = 1$, we have the following convergence rate (up to logarithmic factors) for local SGD (initialized at zero):
    
    \noindent\resizebox{\linewidth}{!}{
    \begin{minipage}{\linewidth}
    \begin{align*}
        &\mathbb{E} \left[F\left(\frac{1}{MKR} \sum_{m=1}^M \sum_{r=1}^{R}\sum_{k=0}^{K-1} x^m_{r,k}\right) - F(x^{\star})\right] \preceq \frac{HB^2}{KR} + \frac{\sigma B}{\sqrt{MKR}} + \frac{(\tau \sigma B^3)^{1/2}}{K^{1/4}R^{1/2}} + \frac{(Q\sigma^2B^5)^{1/3}}{K^{1/3}R^{2/3}} \\ 
        &  + \min\cb{\frac{(\tau \zeta B^3)^{1/2}}{R^{1/2}}  + \frac{(Q\zeta^2B^5)^{1/3}}{R^{2/3}},  \frac{(\tau \zeta_\star B^3)^{1/2}}{R^{1/2}}  + \frac{(Q\zeta_\star^2B^5)^{1/3}}{R^{2/3}} + \frac{(\tau^2DB^3)^{1/2}}{R^{1/2}}  + \frac{(Q\tau^2D^2B^5)^{1/3}}{R^{2/3}} } \,.
    \end{align*}
    \end{minipage}
    }
\end{theorem}
\begin{remark}
    Note that the bound implied by Proposition \ref{prop:zeta_bound} is a worst-case bound for problems in the class $\ppp_{\zeta_\star}^{H,B,\sigma}\cap \ppp_\tau^{H,B,\sigma}$. And in general, there might exist an upper bound $\zeta(\bb_2(D)) \llll \zeta_\star + \tau D$. For this reason, we have stated our upper bounds in terms of $\zeta$ as well as only in terms of $\tau, \zeta_\star, D$. 
\end{remark}
To prove Theorem \ref{col: convex upper bound} we first obtain a convergence rate in the strongly convex setting (Theorem \ref{thm: LSGD_upper_bound}) and then use the standard convex to strongly convex reduction by adding appropriate regularization to our original problem. To the best of our knowledge, Theorem \ref{col: convex upper bound} is the only upper bound for local SGD for problems in the class $\ppp^{H,Q,B,\sigma}_{\tau}~\cap~\ppp^{H,B,\sigma}_{\zeta(\bb_2(D))}$ even for $D=\infty$, i.e., $\xxx=\rr^d$ in Assumption \ref{ass:zeta_everywhere}. Note that the upper bound \textbf{can} benefit from second-order heterogeneity, i.e., Assumption \ref{ass:tau}. In the extreme case, i.e., the homogeneous setting where $\tau, \zeta=0$, our rate recovers the convergence rate by \cite{yuan2020federated} (see Table \ref{tab:summary2}). To compare our result to existing results in the heterogeneous setting by \citet{woodworth2020minibatch}, and to highlight the benefit of Assumptions \ref{ass:tau} and \ref{ass: lipschitz hessian}, we will now compare requirements on $\zeta(\bb_2(D))$ to reach some target sub-optimality $\epsilon$. For simplicity, we will assume that $K$ is large enough to ignore all the terms in the convergence bound with $K$ in the denominator. Then the requirements on $\zeta$ are,
\begin{align}
    \zeta_{old} \preceq \frac{\epsilon^{3/2}R}{H^{1/2}B^2} \qquad\text{ v/s }\qquad \zeta_{our} \preceq \min\cb{\frac{\epsilon^{3/2}R}{(QB)^{1/2}B^2}, \frac{\epsilon^2R}{\tau B^3}}.
\end{align}
In the regime when $Q,\tau$ are small, this highlights that our requirements are much less stringent on $\zeta$. The main limitation of our upper bound, however, is that we assume there exists a $D$ such that, $$\sup_{m\in[M], r\in[R], k\in[0,K-1]}\norm{x_{r,K}^m}\leq D,$$ and prove the upper bound in terms of such a $D$. In fact, we require this for all (random) sequences of local SGD iterates. This is a strong assumption, but we believe it can be avoided by choosing appropriate $\eta, \beta$ in terms of problem-dependent quantities such as $\zeta_\star, \tau, Q$ so that we can control the norm of local SGD iterates (with high probability). We leave this task for future work. For now, we conjecture the following result.
\begin{mdframed}
\begin{conjecture}[Effectiveness of Local SGD]\label{conjecture}
    Local SGD can dominate mini-batch SGD over the problem class $\ppp_{\tau}^{H,Q,B,\sigma}~\cap~\ppp_{\zeta_\star}^{H,B,\sigma}$ in the regime when $\tau, QB \llll H$ and $\sigma, \zeta_\star$ are small.
\end{conjecture}    
\end{mdframed}

\begin{remark}[Limitations of mini-batch SGD in the worst case]
    In our setting, mini-batch SGD can not improve from additional assumptions on top of convexity and smoothness (at least in the worst case). To illustrate this, we assume that all machines share the same objective function $F$, a quadratic function that will be determined later. In this simple setup, note that all data heterogeneity measures are zero; since the objective is quadratic, $Q=0$ as well. Essentially, mini-batch SGD reduces to optimizing $F$ with $R$ mini-batch updates, each with batch size $MK$. Using Lemma \ref{lem:condition_number} and standard sample complexity lower bounds for mean-estimation (also a quadratic problem), we can argue that the convergence rate for mini-batch SGD in this setting is $\frac{HB^2}{R} + \frac{\sigma B}{\sqrt{MKR}}$, which does not benefit from $\tau, \zeta_\star, Q=0$. On the other hand, we conjecture above that local SGD can significantly benefit from these low measures of heterogeneity and sharpness. Thus, if the conjecture is accurate, we would identify a regime of low data heterogeneity where local SGD dominates mini-batch SGD.
\end{remark}

\begin{remark}[Breaking our lower bound]
    Another motivation for our conjecture is the observation that in Example \ref{eq:motivation} as well as our construction for proving Theorem \ref{thm:new_LSGD_lower_bound}, the Hessians on the different machines must be dissimilar. We can not cause the machines to drift between communication rounds if they are not dissimilar, which is, in fact, not possible when $\tau=0$. This is precisely why we believe controlling the distance between the machines' Hessians can break our lower bound. 
\end{remark}

While we do not yet know how to prove Conjecture \ref{conjecture} for the entire class $\ppp_{\tau}^{H,Q,B,\sigma}~\cap~\ppp_{\zeta_\star}^{H,B,\sigma}$, we will prove a very special case of this conjecture in the next section for strongly convex quadratic objectives on each machine which lie in the problem class $\ppp_{\tau}^{H,Q=0,B,\sigma=0}~\cap~\ppp_{\zeta_\star}^{H,B,\sigma=0}$. Quadratic objectives are analytically simpler, and we rely on a fixed-point perspective to understand the behavior of local gradient descent.

\section{Fixed Point of Local SGD for Strongly Convex Objectives}\label{sec:fixed}

Throughout this section, we will assume the objectives of each machine are of the following form:
\begin{align}\label{eq:quadratic_obj}
    F_m(x) = \frac{1}{2}(x-x_m^\star)^TA_m(x-x_m^\star),\ \forall m\in[M] \,,
\end{align}
where $A_m$ has minimum and maximum eigenvalues $\mu,H>0$ respectively and $x_m^\star$ is the unique minimizer of $F_m$. We will denote the optimum of the average objective $x^\star$ and the average of the optima of the machines $\bar x^\star$ as follows where $A:=\frac{1}{M}\sum_{m\in[M]}A_m$ is the average hessian: 
\begin{align}
    x^\star := \frac{1}{M}\sum_{m\in[M]}A^{-1}A_mx_m^\star, \quad \text{ and } \quad \bar x^\star := \frac{1}{M}\sum_{m\in[M]}x_m^\star,    
\end{align}
We will only consider the noiseless setting in this section. Given the quadratic form of the machines' objectives, we can also find a closed-form expression for the fixed point of local gradient descent.
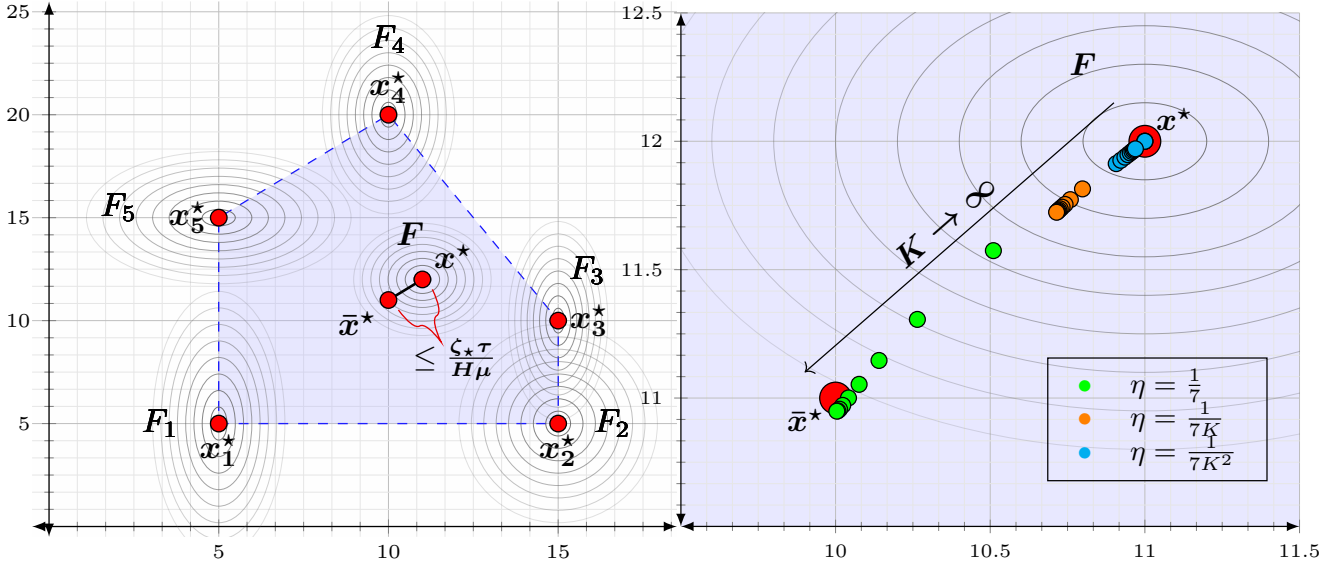
\begin{figure}[ht]
\resizebox{0.9\linewidth}{!}{
\centering
\hspace{-2.8em}
\begin{tikzpicture}
\node at (0,0) {
\begin{minipage}{0.45\textwidth}
\centering
\resizebox{1.3\textwidth}{!}{%

\begin{tikzpicture}
\begin{axis}[
    xmin=0, xmax=18,
    ymin=0, ymax=25,
    grid=both,
    grid style={line width=.1pt, draw=gray!20},
    major grid style={line width=.2pt,draw=gray!50},
    axis lines=middle,
    minor tick num=5,
    enlargelimits={abs=0.5},
    axis line style={latex-latex},
    axis background/.style={fill=gray!1},
    ticklabel style={font=\tiny},
    xlabel style={at={(ticklabel* cs:1)},anchor=north west},
    ylabel style={at={(ticklabel* cs:1)},anchor=south west}
]

\addplot+[only marks, mark=*, mark size=2.5pt, mark options={fill=red, draw=black}]
    coordinates {
    (5,5) (15,5) (15,10) (10,20) (5,15) (10,11) (11,12)
    };
\node at (axis cs:5,5) [anchor=north] {$\bm{x_1^\star}$};
\node at (axis cs:15,5) [anchor=north] {$\bm{x_2^\star}$};
\node at (axis cs:15,10) [anchor=west] {$\bm{x_3^\star}$};
\node at (axis cs:10,20) [anchor=south] {$\bm{x_4^\star}$};
\node at (axis cs:5,15) [anchor=east] {$\bm{x_5^\star}$};

\draw[dashed, blue] (axis cs:5,5) -- (axis cs:15,5) -- (axis cs:15,10) -- (axis cs:10,20) -- (axis cs:5,15) -- cycle;
\filldraw[fill=blue!30, fill opacity=0.3, draw=none] (axis cs:5,5) -- (axis cs:15,5) -- (axis cs:15,10) -- (axis cs:10,20) -- (axis cs:5,15) -- cycle;

\node at (axis cs:10,11) [anchor=north east] {$\bm{\bar x^\star}$};
\node at (axis cs:11,12) [anchor=south west] {$\bm{x^\star}$};

\foreach \r in {1,...,8}{ 
    \pgfmathsetmacro{\step}{\r*2.5}
    \pgfmathsetmacro{\opacity}{max(0.2, 1 - \r / 10)}
    \edef\temp{\noexpand\draw[black!60, very thin, opacity=\opacity] (axis cs:5,5) ellipse ({\step pt} and {\step*2 pt});}
    \temp
    \node[label={[label distance=0cm]180:$F_1$}] at (axis cs:4.5,5) {};
    \edef\temp{\noexpand\draw[black!60, very thin, opacity=\opacity] (axis cs:15,5) ellipse ({\step*1.5 pt} and {\step*1.5 pt});}
    \temp
    \node[label={[label distance=-1.2cm]180:$F_2$}] at (axis cs:14.5,5) {};
    \edef\temp{\noexpand\draw[black!60, very thin, opacity=\opacity] (axis cs:15,10) ellipse ({\step*0.75 pt} and {\step*1.5 pt});}
    \temp
    \node[label={[label distance=0.2cm]45:$F_3$}] at (axis cs:14.25,10) {};
    \edef\temp{\noexpand\draw[black!60, very thin, opacity=\opacity] (axis cs:10,20) ellipse ({\step pt} and {\step*1.5 pt});}
    \temp
    \node[label={[label distance=0.3cm]90:$F_4$}] at (axis cs:10,20.5) {};
    \edef\temp{\noexpand\draw[black!60, very thin, opacity=\opacity] (axis cs:5,15) ellipse ({\step*2 pt} and {\step pt});}
    \temp
    \node[label={[label distance=-1.5cm]0:$F_5$}] at (axis cs:5,15.5) {};
}

\foreach \r in {1,...,8}{
    \pgfmathsetmacro{\stepX}{\r*sqrt(3)} 
    \pgfmathsetmacro{\stepY}{\r*sqrt(2)} 
    \pgfmathsetmacro{\opacity}{max(0.2, 1 - \r / 10)}
    \edef\temp{\noexpand\draw[black!60, very thin, opacity=\opacity] (axis cs:11,12) ellipse ({\stepX*1.5 pt} and {\stepY*1.5 pt});}
    \temp
}
\node[label={[label distance=-0.1cm]150:$\bm{F}$}] at (axis cs:11.5,13) {};

\draw [thick] (axis cs:11,12) -- (axis cs:10,11);
\draw [decorate,decoration={brace,amplitude=15pt},xshift=3pt,yshift=-3pt, color=red!90!black]
(axis cs:11,12) -- (axis cs:10,11) node [black,midway,xshift=12pt,yshift=-18pt] {\small$\leq\bm{\frac{\zeta_\star\tau}{H\mu}}$};

\end{axis}
\end{tikzpicture}
}
\end{minipage}
};
\node at (8.5,0) {
\begin{minipage}{0.45\textwidth}
\centering
\resizebox{1.37\textwidth}{!}{%

\begin{tikzpicture}
\begin{axis}[
    xmin=10, xmax=11,
    ymin=11, ymax=12,
    grid=both,
    grid style={line width=.1pt, draw=gray!20},
    major grid style={line width=.2pt,draw=gray!50},
    axis lines=middle,
    minor tick num=5,
    enlargelimits={abs=0.5},
    axis line style={latex-latex},
    axis background/.style={fill=blue!30, fill opacity=0.3},
    ticklabel style={font=\tiny},
    xlabel style={at={(ticklabel* cs:1)},anchor=north west},
    ylabel style={at={(ticklabel* cs:1)},anchor=south west}
]

\addplot+[only marks, mark=*, mark size=5pt, mark options={fill=red, draw=black}]
    coordinates {
    (10,11) (11,12)
    };
\node at (axis cs:10,11) [anchor=north east] {$\bm{\bar x^\star}$};
\node at (axis cs:11,12) [anchor=south west] {$\bm{x^\star}$};

\addplot+[only marks, mark=*, mark size=2.5pt, mark options={fill=green, draw=black}]
    coordinates {
    (11, 12)
    (10.51020408, 11.57407407)
    (10.26448363, 11.30620985)
    (10.14049587, 11.14633494)
    (10.07609909, 11.05334492)
    (10.04177659, 11.00042072)
    (10.0231446, 10.97136128)
    (10.01290431, 10.95654894)
    (10.00722831, 10.95023809)
    (10.00406319, 10.9489501)
    };

\addplot+[only marks, mark=*, mark size=2.5pt, mark options={fill=orange, draw=black}]
    coordinates {
    (11, 12)
    (10.79831933, 11.81451613)
    (10.75903614, 11.77268722)
    (10.74197075, 11.7540491)
    (10.73240831, 11.74348994)
    (10.72628847, 11.73669031)
    (10.72203465, 11.73194517)
    (10.71890591, 11.72844536)
    (10.71650785, 11.72575743)
    (10.71461121, 11.72362817)
    };

\addplot+[only marks, mark=*, mark size=2.5pt, mark options={fill=cyan, draw=black}]
    coordinates {
    (11, 12)
    (10.90733591, 11.91287879)
    (10.92232661, 11.9259796)
    (10.93551932, 11.93831929)
    (10.94530409, 11.94760228)
    (10.95263775, 11.95459758)
    (10.95828684, 11.95999987)
    (10.96275457, 11.96427829)
    (10.9663694, 11.96774267)
    (10.96935104, 11.97060156)
    };

\foreach \r in {1,...,8}{
    \pgfmathsetmacro{\stepX}{\r*0.2} 
    \pgfmathsetmacro{\stepY}{\r*0.15} 
    \pgfmathsetmacro{\opacity}{max(0.2, 1 - \r / 10)}
    \edef\temp{\noexpand\draw[black!60, very thin, opacity=\opacity] (axis cs:11,12) ellipse ({\stepX} and {\stepY});}
    \temp
}

\node at (axis cs:10.8, 12.3) {$\bm{F}$};

\draw[->] (10.9,12.15) -- (9.9,11.1) node[midway, above, sloped, pos=0.5] {$\bm{K\to\infty}$};

\end{axis}
\node[
    draw,
    at={(6.5,0.5)},
    anchor=south east,
    font=\small
    ] {
    \begin{tabular}{cl}
        \textcolor{green}{$\bullet$} & $\eta = \frac{1}{7}$ \\
        \textcolor{orange}{$\bullet$} & $\eta = \frac{1}{7K}$ \\
        \textcolor{cyan}{$\bullet$} & $\eta = \frac{1}{7K^2}$
    \end{tabular}
};

\end{tikzpicture}
}
\end{minipage}
};
\end{tikzpicture}
}
\caption{Illustration of a two-dimensional optimization problem with $M=5$ machines, each with a $1$-strongly convex, and $6$-smooth objective. On the left figure, we draw the contour lines for each of the machine's objective as well as for the average objective. We also indicate the two relevant solution concepts $\bar x^\star$ and $x^\star$ in the same figure, noting that their distance is bounded by Proposition \ref{prop:bar_star_distance}. On the right figure, we zoom into the convex hull of the machines' optima, noting the sequence of fixed points for local GD as a function of $\eta$ and increasing $K\in[10]$. We plot the fixed points for three different choices of $\eta$ each demonstrating a different trend for $\lim_{K\to\infty}x_{\infty}(K, \eta, \beta)$.}
\label{fig:fixed}
\end{figure}

\begin{proposition}\label{prop:lsgd_fixed_point}
    Consider an instance of the problem of the form \eqref{eq:quadratic_obj} with strongly convex and smooth machines' objectives. Let $x_{R}\rb{K, \eta, \beta}$ be the local gradient descent iterate after $R$ communication rounds, using exact gradient calls (i.e., $\sigma=0$), $K$ local updates, and step-sizes, $\eta, \beta$. Then the fixed point for local GD (whenever it exists) is given by,
    \begin{align*}
        x_{\infty}\rb{K, \eta, \beta} := \lim_{R\to\infty} x_{R}\rb{K, \eta, \beta} = \frac{1}{M}\sum_{m\in[M]}C^{-1}C_m x_m^\star,
    \end{align*}
    where $C_m = I- \rb{I-\eta A_m}^K$ and $C=\frac{1}{M}\sum_{m\in[M]}C_m$. 
\end{proposition}
\begin{remark}
    Note that the fixed point might not be defined for all choices of $\eta, \beta, K$. Intuitively, it should be well defined when $\eta, \beta$ are \textit{``small enough"} relative to how $R$ increases. Also note that since $\sigma=0$, the fixed point is non-random. When $\sigma>0$, we instead need to look at the expected local SGD iterate. 
\end{remark}
\begin{figure}
    \centering
    \includegraphics[width=0.7\textwidth]{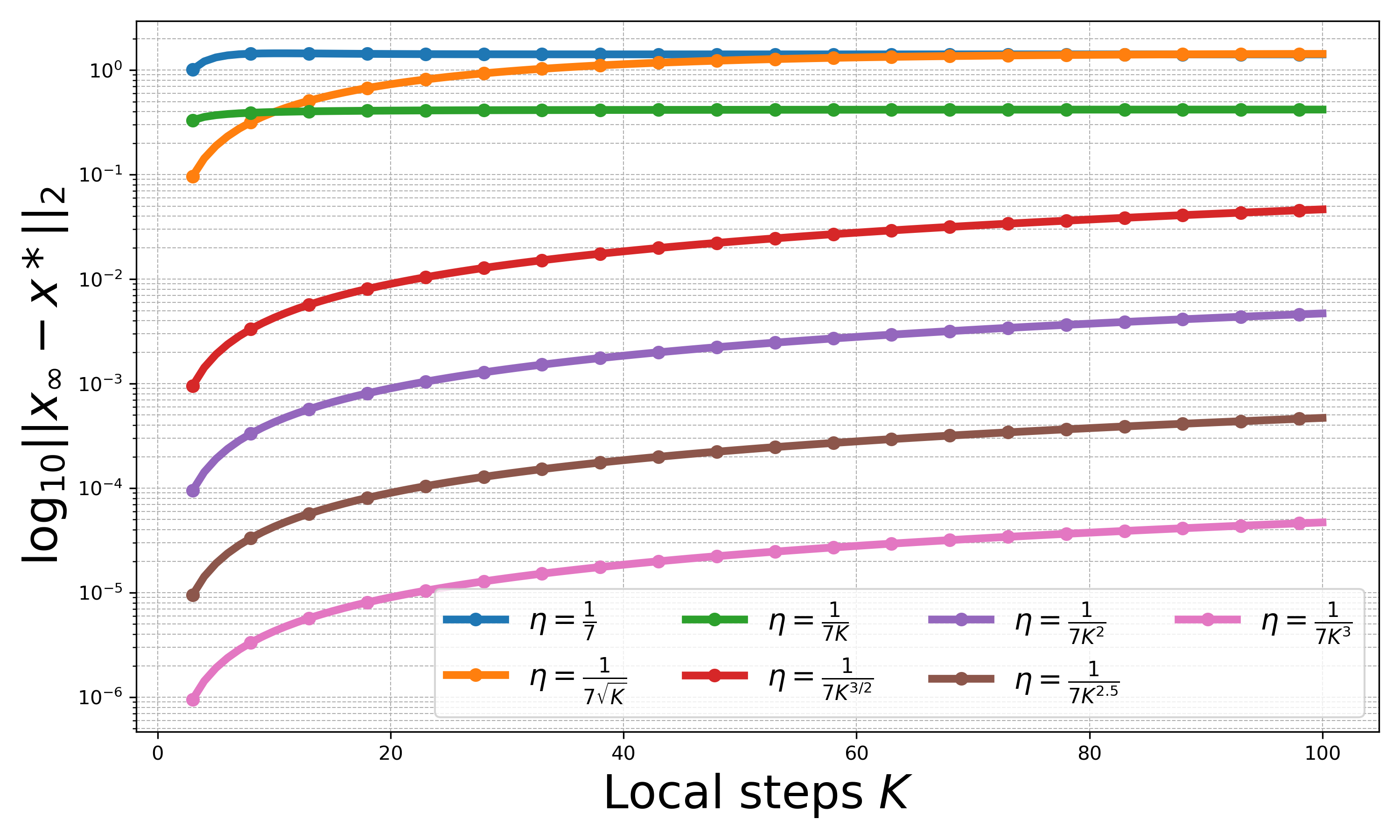}
    \caption{Illustration of the same distributed problem as Figure \ref{fig:fixed} to understand the where the fixed point converges as $K$ grows. We consider $7$ different choices of $\eta$ (as a function of $K$) and plot $\log\norm{x_{\infty}(K,\eta,1)-x^\star}$ as a function of $K\in[100]$. We notice that for $\eta >\frac{1}{HK}$, the fixed point goes to $\bar x^\star$ as $K$ increases, while for $\eta<\frac{1}{HK}$, the fixed point gets progressively closer to $x^\star$. }
    \label{fig:convergence}
\end{figure}

\noindent Using the closed form of $x_\infty$ from Proposition \ref{prop:lsgd_fixed_point} we can make two observations (c.f., Figure \ref{fig:fixed}):
\paragraph{1. The correct fixed point.} When $K=1$, for any choice of $\eta, \beta >0$ that ensures convergence of local GD, $x_\infty=x^\star$. This is unsurprising because with $K=1$, local GD reduces to GD on the averaged objective, so the fixed point must be a stationary point of $F$, i.e., $x^\star$. We also observe in Figure \ref{fig:convergence} that as $\eta$ decreases as a function of $K$, and $K\to\infty$ we recover the correct fixed point. This is also expected, as the effect of local updates diminishes with smaller $\eta$.

\paragraph{2. The incorrect fixed point.} On the other hand when $K\to\infty$ and we pick $\eta=\omega\rb{\frac{1}{KH}}$ but $\eta<\frac{1}{H}$ we get that $\lim_{K\to\infty}x_{\infty}(K,\eta, \beta) = \bar{x}^\star$. This makes intuitive sense (at least for $\beta=1$) because for a large step-size $\eta$, that also ensures convergence of GD on each machine, local GD between communication rounds would converge to machines' optima, and then averaging the optima would yield $\bar x^\star$.   
In fact, using Assumptions \ref{ass:tau} and \ref{ass:zeta_optimal_stronger} (a strongly convex variant of Assumption \ref{ass:zeta_optimal} on $S^\star$), we can show that the distances between $\bar x^\star$, $x^\star$ and $x_\infty$ are bounded. Specifically, we consider the following assumption,
\begin{assumption}[Bounded Optima Heterogeneity]\label{ass:zeta_optimal_stronger}  
    A set of $H$-smooth and strongly convex objectives $\{F_m\}_{m\in[M]}$, satisfy $\zeta_\star$-optima-heterogeneity if,
    $$\frac{1}{M}\sum_{m\in[M]}\norm{x_m^\star - x^\star}^2 \leq \frac{\zeta_\star^2}{H^2},$$
    where $x_m^\star$ is the optimum of $F_m$ for all $m\in[M]$ and $x^\star\in S^\star$ is the optimum of the average objective $F$.
\end{assumption}
Note that using smoothness along with Assumption \ref{ass:zeta_optimal_stronger} implies Assumption \ref{ass:zeta_everywhere} for $\xxx =S^\star$. Intuitively, when the machines have unique optima, the most natural measure of heterogeneity is the distance between these optima and the optima of the average objective, which the above assumption measures\footnote{A problem instance with objectives of the form \eqref{eq:quadratic_obj} satisfying Assumption \ref{ass:zeta_optimal_stronger} lies in the problem class $\ppp_{\zeta_\star}^{H,B,\sigma}$.}. This assumption enables us to prove the following upper bound on the fixed-point discrepancy of local gradient descent.
\begin{proposition}\label{prop:bar_star_distance}
    Consider a problem instance of the form \eqref{eq:quadratic_obj} with $\mu$-strongly convex and $H$-smooth machines' objectives satisfying Assumptions \ref{ass:tau} and \ref{ass:zeta_optimal_stronger}, then for $K>1$,
    \begin{align*}
        \norm{x^\star - \bar x^\star} \leq \frac{\zeta_\star \tau}{H \mu}\quad \text{ and }\quad \norm{x_{\infty}(K,\eta, \beta) - \bar x^\star} \leq \frac{\zeta_\star \tau}{H\mu}\cdot \frac{\eta \mu K\rb{1-\eta \mu}^{K-1}}{1-(1-\eta\mu)^K}\preceq \enspace \frac{\zeta_\star \tau}{H\mu}.
    \end{align*}
\end{proposition}
Using the triangle inequality, Proposition \ref{prop:bar_star_distance} bounds the fixed point discrepancy for local SGD by some constant times $\frac{\zeta_\star \tau}{H\mu}$. Combining this with a convergence guarantee for local SGD to its fixed point, we obtain the following convergence guarantee in the noiseless setting (compare to \citet{wang2022unreasonable} in Appendix \ref{app:wang}). 
\begin{theorem}[Local GD with Strongly Convex Objectives]\label{thm:fixed_convergence}
Consider a problem instance of the form \eqref{eq:quadratic_obj} with $\mu$-strongly convex and $H$-smooth machines' objectives satisfying Assumptions \ref{ass:tau} and \ref{ass:zeta_optimal_stronger}, then for $\beta = 1$ and $\eta=\frac{1}{2H}$ we get the following convergence for local GD initialized at zero,
    \begin{align*}
        \norm{x_R-x^\star} &\preceq e^{-\frac{KR}{\kappa}}\cdot \frac{1-\rb{\frac{1}{2}}^K}{1-\rb{1-\frac{1}{2\kappa}}^K}B + \frac{\zeta_\star \tau}{H \mu}.
    \end{align*}
Assuming $K\geq \frac{1}{-\log_2\rb{1-\frac{1}{2\kappa}}}$ this simplifies to,
\begin{align*}
        \norm{x_R-x^\star} &\preceq Be^{-\frac{KR}{\kappa}} + \frac{\zeta_\star \tau}{H \mu}.
    \end{align*}
\end{theorem}
Comparing this bound to the rate for GD (i.e., the noiseless variant of mini-batch SGD), we see that when $\zeta_\star, \tau$ are small, local GD has a much better optimization term v/s the optimization term $Be^{-R/\kappa}$ of GD. This implies that we have identified a $\mu$-strongly sub-class of  $\ppp_{\tau}^{H,Q,B,\sigma=0}~\cap~\ppp_{\zeta_\star}^{H,B,\sigma=0}$, where local GD dominates GD. This further motivates us to believe that Conjecture \ref{conjecture} is correct. We note that the analyses characterizing the fixed-point of local SGD have also been considered by \citet{malinovskiy2020local} and \citet{charles2020outsized}. However, in the above discussion, we hope to underline how this view relates to our study of data heterogeneity notions.

\subsection{Two Stage Algorithms}
    Based on the discussion in this section, a natural question is how to reduce the fixed-point discrepancy of local SGD. One approach based on the observation in Figure \ref{fig:fixed} is to reduce the inner step-size of local SGD as iterations proceed so that we get the benefit of more aggressive updates at the beginning of training but can also avoid the fixed-point discrepancy at the end of training. Note that reducing the inner step size actually reduces local SGD to mini-batch SGD in the limit. Thus, an even easier approach to fixing the fixed point issue is sharply transitioning from running local SGD to mini-batch SGD (which has no fixed point issues) at the end of training. 
    
    Such a two-stage algorithm has been considered by \citet{hou2021fedchain}, motivated by the results of \citet{woodworth2020minibatch}. Theorem \ref{thm:fixed_convergence} provides a more fine-grained perspective on this two-stage algorithm by capturing the effect of $\zeta_\star, \tau$ as opposed to $\zeta(\rr^d)$ in \citet{hou2021fedchain}. In particular, we run local GD for the first phase until the optimization (first) term in Theorem \ref{thm:fixed_convergence} dominates the convergence, i.e., $R_1 = \theta \rb{\frac{\kappa}{K}\ln\rb{\frac{H\mu}{\zeta_\star \tau B}}}$ rounds. At this point, we switch to running GD, i.e., the noiseless variant of mini-batch SGD, for $R_2 = \theta\rb{\kappa\ln\rb{\frac{\zeta_\star \tau}{H\mu\epsilon}}}$, for some target accuracy $\epsilon$. Thus, in total, the communication complexity of this algorithm is given by, $$R(\epsilon) = R_1 + R_2 = \theta\rb{\kappa\ln\rb{\rb{\frac{H\mu}{\zeta_\star \tau}\cdot \frac{1}{B}}^{1/K}\rb{\frac{\zeta_\star \tau}{H\mu}\cdot \frac{1}{\epsilon}}}}.$$     
This communication complexity exemplifies the effect of local update steps, where we notice that in the limit $K\to\infty$ the communication complexity of the two-stage algorithm local GD->GD is much better than running GD alone as long as $\frac{\zeta_\star \tau }{\mu} << HB$.

\subsection{Interpreting the Heterogeneity Assumptions}
So far, we have introduced several heterogeneity assumptions in this paper. We end this discussion by considering a simple linear regression model to highlight how these assumptions might appear in real-world problems relating to actual distribution shifts between clients. We will consider a linear regression problem, where each data point is a co-variate label pair $z_m := (\beta_m, y_m)\sim \ddd_m$ on each client such that:
\begin{itemize}
    \item The covariates are sampled from a Gaussian distribution, i.e., $\beta_m\sim \nnn(\mu_m, \Sigma_m)\in\rr^d$; and
    \item The labels are generated as $y_m \sim \inner{x_m^\star}{\beta_m} + \nnn(0, \sigma^2_{noise})\in\rr$ for some ground truth model $x_m^\star\in \rr^d$ on machine $m$.
\end{itemize}
The goal of each client is to minimize the mean squared error, i.e., $f(x;(\beta_m, y_m)) = \frac{1}{2}\rb{y_m - \inner{x}{\beta_m}}^2$ and the expected loss function is given by, 
\begin{align*}
  F_m(x) &= \ee_{z_m\sim \ddd_m}[f(x;z_m)]\enspace,\\
  &= \ee_{\beta_m\sim \nnn(\mu_m, \Sigma_m), \xi_m\sim \nnn(0, \sigma^2_{noise})}\sb{\frac{1}{2}\rb{\inner{x_m^\star - x}{\beta_m} + \xi_m}^2}\enspace,\\
  &= \ee_{\beta_m\sim \nnn(\mu_m, \Sigma_m)}\sb{\frac{1}{2}\rb{\inner{x_m^\star - x}{\beta_m}}^2} + \ee_{\xi_m\sim \nnn(0, \sigma^2_{noise})}\sb{\frac{1}{2}\xi_m^2}\enspace,\\
  &= \frac{1}{2}(x-x_m^\star)^T\ee_{\beta_m\sim \nnn(\mu_m, \Sigma_m)}\sb{\beta_m\beta_m^T}(x-x_m^\star) + \frac{1}{2}\sigma_{noise}^2\enspace,\\
  &= \frac{1}{2}(x-x_m^\star)^T\rb{\mu_m\mu_m^T + \Sigma_m}(x-x_m^\star) + \frac{1}{2}\sigma_{noise}^2\enspace,
\end{align*}
which fits in the model of $H$-smooth and strongly convex quadratic functions we have considered in this section, as long as $0\prec \mu_m\mu_m^T + \Sigma_m \preceq H.I_d$. Interestingly, Assumptions \ref{ass:tau} and \ref{ass:zeta_optimal_stronger} have a very nice interpretation in this setting, 
\begin{itemize}
    \item Note that for any $m,n\in[M]$, $$\norm{A_m - A_n} \leq (\norm{\mu_m} + \norm{\mu_n})\norm{\mu_m-\mu_n} + \norm{\Sigma_m - \Sigma_n}=:\tau_{m,n},$$
    Thus this problem satisfies Assumption \ref{ass:tau} with $\tau^2:= \sup_{m,n}\tau^2_{m,n}$ which in turn can be thought of as a measure of the \textit{\textbf{``co-variate shift"}} between the machines.
    \item Similarly, note that for any $m,n\in[M]$, $$\norm{x_m^\star - x_n^\star}^2 \leq 2\norm{x_m^\star - x^\star}^2 + 2\norm{x_n^\star - x^\star}^2,$$
    for $x\in S^\star$. Thus Assumption \ref{ass:zeta_optimal_stronger} controls the average \textit{\textbf{``concept-shift"}} between the clients as,
    $$\frac{1}{M^2}\sum_{m,n\in[M]}\norm{x_m^\star - x_n^\star}^2 \leq \frac{4}{M}\sum_{m\in[M]}\norm{x_m^\star - x^\star}^2 \leq 4\frac{\zeta_\star^2}{H^2}.$$
\end{itemize}
In light of this interpretation, note that Theorem \ref{thm:fixed_convergence} implies that the local SGD can approach a fast convergence rate, i.e., not suffer due to the fixed point discrepancy, as long as either the co-variate shift ($\tau$) or the concept shift ($\zeta_\star$) is low. We believe this provides a useful interpretation for these heterogeneity assumptions, and this simple learning problem might pave the way to a better understanding of distributed learning algorithms. In particular, we posit the following problem, which is largely open: 
\begin{mdframed}
\begin{oproblem}
    For the distributed linear regression problem described in this setting, and for different regimes of heterogeneity/distribution shift parameters $\tau, \zeta_\star$ as well as problem dependent parameters $H, \mu, \sigma, B$ identifying the min-max optimal algorithms. 
\end{oproblem}    
\end{mdframed}

Answering the above problem requires analyzing other algorithms, such as the two-stage algorithm mentioned in the previous sub-section, reconciling existing literature in low heterogeneity regimes, including the homogeneous regime, and identifying relevant lower bounds for the problem. Most importantly, the insights from answering the above question might have many useful prescriptions for the practice of distributed/federated learning in the face of different natures of heterogeneity, such as co-variate/concept shift. We hope to address this problem more comprehensively in future work.


\section{Discussion}\label{sec:discuss}
In this paper, we further the discussion around the convergence rate of local SGD and establish the min-max optimal convergence rate for the much-studied problem class $\ppp^{H, B, \sigma}_{\zeta_\star}$. While the lower bound results of our paper seem to paint a negative picture for local SGD, we believe their real message is to help us identify the additional heterogeneity assumptions required to demonstrate the effectiveness of local SGD. Based on the paper's upper bound results and our current insights, we believe that combining first-order heterogeneity assumptions with higher-order assumptions is key to understanding the ``real" behavior of local SGD. Towards this end, we believe the central message of our paper is Conjecture \ref{conjecture}, stated in Section \ref{sec:lower}.
Proving this conjecture is an important future direction, and we believe it can both enhance our understanding of local SGD and aid in designing a better algorithm for practice. 

\paragraph{Acknowledgemnts.} We thank the anonymous reviewers at COLT, ICML, and NeurIPS, who helped significantly improve the paper. We also thank Blake Woodworth and Brian Bullins for several useful discussions. KKP was supported through the NSF TRIPOD Institute on Data, Economics, Algorithms and Learning (IDEAL) and other awards from DARPA and NSF. AZ and SUS acknowledge partial funding from a Google Scholar Research Award.

\bibliographystyle{unsrtnat}
\bibliography{references}  

\newpage
\appendix


\section{Missing Details from Section \ref{sec:setting}}\label{app:setting}

\subsection{Another First-order Heterogeneity Assumption} 
\citet{karimireddy2020scaffold} looked at a following relaxed first-order heterogeneity assumption, but they were \textbf{not} able to show that local SGD dominates large mini-batch SGD under this assumption.
\begin{assumption}[Relaxed First-Order Heterogeneity Everywhere]\label{ass:relaxed_zeta_everywhere}
    A set of objectives $\{F_m\}_{m\in[M]}$ satisfy $(\zeta_\star,D)$-first-order heterogeneity everywhere if for all $x\in \rr^d$, 
    $$\frac{1}{M}\sum_{m\in[M]}\norm{\nabla F_m(x)}^2 \leq \zeta_\star^2 + D^2\norm{\nabla F(x)}^2.$$
\end{assumption}
This assumption \textit{``interpolates"} between Assumption \ref{ass:zeta_everywhere} over the unbounded domain $\xxx=\rr^d$ v/s at the optima $\xxx=S^\star$ for different values of $D$. Furthermore, the restrictiveness of Assumption \ref{ass:zeta_everywhere} pointed out in Proposition \ref{prop:zeta_limitation} does not extend to this assumption as long as $D>0$. While we did not look at this assumption in this paper, it is also promising for future work. 

\subsection{Proof of Proposition \ref{prop:zeta_limitation}}
\begin{proof}
    Note the following for any $m\in[M]$ using triangle inequality,
    \begin{align*}
        sup_{x\in \rr^d}\norm{\nabla F_m(x) - \nabla F(x)} &= sup_{x\in \rr^d}\norm{(A_m-A)x + b_m-b},\\
        &\geq sup_{x\in \rr^d}\norm{(A_m-A)x} - \norm{b_m-b}.
    \end{align*}
Denote the matrix $C_m:= A_m-A = [c_{m,1}, \dots, c_{m,d}]$ using its column vectors. Then take $x = \delta e_i$ where $e_i$ is the $i$-th standard basis vector to note in the above inequality,
\begin{align*}
        sup_{x\in \rr^d}\norm{\nabla F_m(x) - \nabla F(x)} &\geq \delta \norm{(A_m-A)e_i} - \norm{b_m-b},\\
        &\geq \delta \norm{c_{m,i}} - \norm{b_m}-\norm{b}.
    \end{align*}
    Assuming $\norm{b_m}, \norm{b}$ are finite, since we can take $\delta\to \infty$ we must have $\norm{c_{m,i}}=0$ for all $i\in[d]$ if $\zeta < \infty$. This implies that $c_{m,i} = 0$ for all $i\in[d]$, or in other words $A_m = A$. Since this is true for all $m\in[M]$, the machines must have the same Hessians, and thus they can only differ upto linear terms.
\end{proof}

\section{Missing Details from Section \ref{sec:lower}}\label{app:lower}
\subsection{Proof of Equation \ref{eq:motivation}}
\begin{proof}
    Note that the update on machine $m$ leading up to communication round $r$ is as follows for $k\in[0,K-1]$ and $m=1$,
    \begin{align*}
     &x_{r,k+1}^1[1] = x_{r,k}^1[1] - \eta H(x_{r,k}^1[1]-x^\star[1]),\\
     &\Rightarrow x_{r,k+1}^1[1] = x^\star[1] + (1-\eta H)^{k+1}(x_{r,0}^1[1]-x^\star[1]),\\
     &\Rightarrow x_{r,K}^1[1] = x^\star[1] + (1-\eta H)^{K}(x_{r-1}[1]-x^\star[1]),\\
     &\Rightarrow x_{r,K}^1[1] - x_{r-1}[1] = (1-(1-\eta H)^{K})(x^\star[1]-x_{r-1}[1]).
    \end{align*}
    On the second dimension, the iterates don't move at all for $m=1$, 
    \begin{align*}
        x_{r,K}^1[2] - x_{r-1}[2] = 0.
    \end{align*}
    Writing a similar expression for the second machine and averaging these updates we get,
    \begin{align*}
        \frac{1}{2}\sum_{m\in[2]}(x_{r,K}^m-x_{r-1}) = \frac{1}{2}(1-(1-\eta H)^{K})(x^\star-x_{r-1}).
    \end{align*}
    This gives the update for communication round $r$ as follows,
    \begin{align*}
        &x_r = x_{r-1} + \frac{\beta}{2}(1-(1-\eta H)^{K})(x^\star-x_{r-1}),\\
        &\Rightarrow x_r - x^\star = \rb{1 - \frac{\beta}{2}(1-(1-\eta H)^{K})}(x_{r-1}-x^\star),\\
        &\Rightarrow x_R = x^\star + \rb{1 - \frac{\beta}{2}(1-(1-\eta H)^{K})}^R(x_{0}-x^\star),\\
        &\Rightarrow x_R = \rb{1 - \rb{1 - \frac{\beta}{2}(1-(1-\eta H)^{K})}^R}x^\star,
    \end{align*}
    which finishes the proof.
\end{proof}

\subsection{Proof of Lemma \ref{lem:condition_number}}
\begin{proof}
Let $A$ be the Hessian of $F$. Observe that we have $F(x) - F(x^\star) = \frac{1}{2}(x - x^\star)^TA(x - x^\star).$

Let $v_1$ and $v_2$ be the eigenvectors of norm $1$ of $A$ with the greatest and least eigenvalues, respectively. Assume $x^\star := -B\left(\frac{v_1 + v_2}{\sqrt{2}}\right)$, which ensures $\|x^\star\|_2 = B$. Then, solving for the GD iterates in closed form, we have 
\begin{equation}
    x_R - x^* = \frac{B}{\sqrt{2}}\left(1 - \eta H\right)^Rv_1 + \frac{B}{\sqrt{2}}\left(1 - \eta \frac{H}{\kappa}\right)^Rv_2. 
\end{equation}
Observe that if $\eta \geq \frac{3}{H}$, then the iterates explode and we have $F(x_R) \geq F(x_0) \geq \Omega\left(HB^2\right)$.

If $\eta \leq \frac{3}{H}$, then using the fact that $\kappa \geq 6$, we have
\begin{align}
    F(x_R) - F(x^\star) &\geq \frac{1}{2}\left(\frac{B}{\sqrt{2}}\left(1 - \frac{3}{\kappa}\right)^Rv_2\right)^T A\left(\frac{B}{\sqrt{2}}\left(1 - \frac{3}{\kappa}\right)^Rv_2\right)\\
    &= \frac{B^2}{4}\left(1 -  \frac{3}{\kappa}\right)^{2R}v_2^TAv_2 \\
    &= \frac{B^2}{4}\left(1 - \frac{3}{\kappa}\right)^{2R}\frac{H}{\kappa}\\
    &\geq \frac{HB^2}{4\kappa} e^{-12R/\kappa}.
\end{align}
The result follows.
\end{proof}

\subsection{Proof of Proposition \ref{prop:lb_zeta0}}
\begin{proof}
    We will consider a two-dimensional problem in the noiseless setting for this proof, as we do not want to understand the dependence on $\sigma$ or $d$. Define $A_1 := \left(\begin{array}{cc}
            1 & 0 \\
            0 & 0
        \end{array}\right)$ and $A_2:=vv^T$, where $v=(\alpha,\sqrt{1-\alpha^2})$ and $\alpha\in (0,1)$. For even $m$, let 
    \begin{equation*}
        F_m(x):= \frac{H}{2}(x-x^*)^TA_1(x-x^*)\enspace.
    \end{equation*}
    For odd $m$, let 
    \begin{equation*}
        F_m(x):= \frac{H}{2}(x-x^*)^TA_2(x-x^*)\enspace.
    \end{equation*}
    Note that $A_1$ and $A_2$ are rank-1 and have eigenvalues $0$ and $1$, and thus they satisfy Assumption \ref{ass:smth}. Furthermore, both the functions have a shared optimizer $x^\star$. It is easy to verify that,
    \begin{equation*}
        (I-\eta HA_i)^K = I-(1-(1-\eta H)^K)A_i=: I-\widetilde{\eta}HA_i,
    \end{equation*}
    where $\widetilde{\eta}:=(1-(1-\eta H)^K) / H$. Note that the above property will be crucial for our construction, and we can not satisfy this property if our matrices are not ranked one. For any $x$, let us denote the centered iterate by $\tilde{x} := x - x^\star$. Then, for any $r\in[R]$  and $m\in[M]$ we have
    \begin{equation*}
        \tilde{x}^{m}_{r,K} = (I-\eta HA_m)^K \tilde{x}_{r-1} = (I-\widetilde{\eta}HA_m)\tilde{x}_{r-1}.
    \end{equation*}
    Using this, we can write the updates between two communication rounds as,
    \begin{align*}
        \tilde{x}_{r} &= \tilde{x}_{r-1} + \frac{\beta}{M}\sum_{m\in[M]}\rb{\tilde{x}_{r,K}^m - \tilde{x}_{r-1}},\\
        &= \tilde{x}_{r-1} - \frac{\beta}{M}\sum_{m\in[M]}\widetilde{\eta}HA_m\tilde{x}_{r-1},\\
        &= \rb{I - \beta \widetilde{\eta}H A}\tilde{x}_{r-1},\\
        &= \tilde{x}_{r-1} - \beta \widetilde{\eta}\nabla F(x_{r-1}),
    \end{align*}
    where we used that $F(x)=\frac{H}{2}(x-x^\star)^TA(x-x^\star)$, for $A = (1-a)A_1 + aA_2$ and 
    \begin{equation*}
        a := \left\{
            \begin{array}{ll}
                1/2 &  \text{if $M$ is even}, \\
                (M+1)/2M & \text{otherwise}.
            \end{array}
        \right.
    \end{equation*}
    This implies the iterates of local GD across communication rounds are equivalent to GD on $F(x)$ with step size $\beta (1-(1-\eta H)^K) / H$. Combining this observation with Lemma \ref{lem:condition_number} about the function sub-optimality of gradient descent updates will finish the proof. To use the lemma, however, we need to verify our average function $F$ has condition number $\Omega(R)$. We can explicitly compute the eigenvalues of $A$ as follows,
    \begin{align*}
        \lambda_1 &= \frac{1}{2} + \sqrt{\frac{1}{4}-(a-a^2)(1-\alpha^2)},\\ 
        \lambda_2 &= \frac{1}{2} - \sqrt{\frac{1}{4}-(a-a^2)(1-\alpha^2)}.
    \end{align*}
    Note that $\lim_{\alpha \rightarrow 1} \lambda_1 = 1$, and  $\lim_{\alpha \rightarrow 1} \lambda_2 = 0$ and thus $\lim_{\alpha \rightarrow 1} \lambda_1/\lambda_2 = \infty$. Since $\lambda_1/\lambda_2=1$ when $\alpha=0$, by the intermediate value theorem, we can choose $\alpha$ to get $\kappa=\Omega(R)$ for the average objective $F$. Thus, we can use Lemma \ref{lem:condition_number} and finish the proof.
\end{proof}    
Note that in the construction used in the above proof, the machines share an optimum point $x^\star$, which ensures that $\zeta_\star=0$.

\subsection{Proof of Theorem \ref{thm:new_LSGD_lower_bound}}
\begin{proof}
    To combine the hard instance in Proposition \ref{prop:lb_zeta0} with the previous hard instance of \cite{glasgow2022sharp} we simply place the two instances on disjoint co-ordinates increasing the dimensionality of our hard instance. This is a standard technique to combine lower bounds. To get rid of the terms with the minimum function in the lower bound of \cite{glasgow2022sharp}, we note the following,
    \begin{itemize}
        \item $\frac{\sigma B}{\sqrt{KR}} \geq \frac{(H\sigma^2B^4)^{1/3}}{K^{1/3}R^{2/3}}$ implies that $\frac{\sigma B}{\sqrt{KR}} \leq \frac{HB^2}{R}$, and
        \item $\frac{\zeta_\star^2}{H} \geq \frac{(H\zeta_\star^2B^4)^{1/3}}{R^{2/3}}$ implies that $\frac{\zeta_\star^2}{H} \leq \frac{HB^2}{R}$.
    \end{itemize}
    These observations allow us to avoid the minimum operations, thus concluding the proof.
\end{proof}


\subsection{Distributed Zero-respecting Algorithms}\label{app:zero_respecting}
Intermittent communication is motivated by the sizeable gap between the wall-clock time $\ccc$ required for a single synchronous communication and the time required per unit of computation $\ttt$, say a single oracle call \citep{mcmahan2016communication, kairouz2019advances}. For an efficient implementation, typically, we want our local computation budget $K$ to be comparable to $\ccc/\ttt$, i.e., we want to increase our computation load per communication to match the time required for a single communication round.  
We consider a generalization of zero respecting algorithms \citep{carmon2020lower} denoted by $\aaa_{ZR}$ in the intermittent communication (IC) setting defined as follows.
\begin{definition}[Distributed Zero-respecting Algorithms]\label{def:zero_respecting}
Consider $M$ machines in the IC setting, each endowed with an oracle $\ooo_m:\iii\times \zzz\to \vvv$ and a distribution $\ddd_m$ on $\zzz$. Let $I_{r,k}^m$ denote the input to the $k^{th}$ oracle call, leading up to the $r^{th}$ communication round on machine $m$. An optimization algorithm initialized at $0$ is distributed zero-respecting if:
\begin{enumerate}
    \item for all $\ r\in [R], k\in[K], m\in[M]$, $I_{r,k}^{m}$ is in
$$\cb{\bigcup_{l\in[k-1]}\supp{\ooo_{F_m}(I_{r,l}^m; z_{r,l}^m\sim \ddd_m)}}\cup \cb{\bigcup_{n\in[M], s\in[r-1], l \in [K]} \supp{\ooo_{F_n}(I_{s,l}^n; z_{s,l}^n\sim \ddd_n)}},$$
    \item for all $\ r\in [R], k\in[K], m\in[M]$, $I_{r,k}^{m}$ is a deterministic function (which is same across all the machines) of $$\cb{\bigcup_{l\in[k-1]}\ooo_{F_m}(I_{r,l}^m; z_{r,l}^m\sim \ddd_m)}\cup \cb{\bigcup_{n\in[M], s\in[r-1], l \in [K]} \ooo_{F_n}(I_{s,l}^n; z_{s,l}^n\sim \ddd_n)},$$
    \item at the $r^{th}$ communication round, the machines only communicate vectors in
$$\cb{\bigcup_{n\in[M], s\in[r], l \in [K]} \supp{\ooo_{F_n}(I_{s,l}^n; z_{s,l}^n\sim \ddd_n)}},$$
\end{enumerate}
where $\supp{.}$ denotes the co-ordinate support/span of its arguments. We denote this class of algorithms by $\boldsymbol{\aaa_{ZR}}$. Furthermore, if all the oracle inputs are the same between two communication rounds, i.e., $I_{r,k}^m = I_r\in \iii$ for all $m\in[M], k\in[K], r\in[R]$, then we say that the algorithm is centralized, and denote this class of algorithms by $\boldsymbol{\aaa_{ZR}^{cent}}\subset \aaa_{ZR}$. 
\end{definition}

This class captures a very wide variety of distributed optimization algorithms, including mini-batch SGD \citep{dekel2012optimal}, accelerated mini-batch SGD \citep{ghadimi2012optimal}, local SGD \citep{mcmahan2016communication}, as well as all the variance-reduced algorithms \citep{karimireddy2020mime, zhao2021fedpage, khanduri2021stem}. Non-distributed zero-respecting algorithms are those whose iterates have components in directions about which the algorithm has no information, meaning that, in some sense, it is just ``wild guessing''. We have also defined the smaller class of centralized algorithms $\aaa_{ZR}^{cent}$. These algorithms query the oracles at the same point within each communication round and use the combined $MK$ oracle queries each round to get a \textit{``mini-batch''} estimate of the gradient. Thus, the class $\aaa_{ZR}^{cent}$ includes algorithms such as mini-batch SGD \citep{dekel2012optimal, woodworth2020minibatch}, but doesn't include local-update algorithms in $\aaa_{ZR}$ such as local-SGD. 

\subsection{Proof of Theorem~\ref{thm:AIlb_zeta0}}
For  even $m$, let 
\begin{equation}
    F_m(x) := \frac{H}{2}\left((q^2 + 1)(q - x_1)^2 + \sum_{i = 1}^{\lfloor{(d - 1)/2}\rfloor} (qx_{2i} - x_{2i + 1})^2\right),
\end{equation}
and for odd $m$, let 
\begin{equation}
    F_m(x) =  \frac{H}{2}\left(\sum_{i = 1}^{\lfloor{d/2}\rfloor} (qx_{2i - 1} - x_{2i})^2\right).
\end{equation}
Thus we have
\begin{equation}
    F(x) = \mathbb{E}_m[F_m(x)] = \frac{H}{2}\left((q^2 + 1)(q - x_1)^2 + \sum_{i = 1}^{\d} (qx_{i} - x_{i + 1})^2\right).
\end{equation}

Observe that the optimum of $F$ is attained at $x^\star$, where $x^\star_i = q^i$. 
Theorem~\ref{thm:AIlb_zeta0} improves on the previous best lower bounds by introducing the term $\frac{HB^2}{R^2}$. Combining the following lemma with standard arguments to achieve the $\frac{\sigma B}{\sqrt{MKR}}$ suffices to prove Theorem~\ref{thm:AIlb_zeta0}.

\begin{lemma}\label{lem:AI}
For any $K \geq 2, R, M, H, B, \sigma$, there exist $f(x; xi)$ and distributions $\{\mathcal{D}_m\}$, each satisfying assumptions~\ref{ass:smth}, \ref{ass:bounded_optima}, \ref{ass:stoch_first_order}, and together satisfying $\frac{1}{M} \sum_{m=1}^M \|\nabla F_m(x^{\star})\|_2^2 = 0$, such that for initialization $x^{(0, 0)}=0$, the final iterate $\hat{x}$ of any zero-respecting with $R$ rounds of communication and $KR$ gradient computations per machine satisfies
\begin{align}
 \mathbb{E}\left[F(\hat{x})\right] - F(x^{\star}) \succeq \frac{HB^2}{R^2}.
\end{align}
\end{lemma}
\begin{proof}
    Consider the division of functions onto machines described above for some sufficiently large $d$.

    Let $q = 1 - \frac{1}{R}$, and let $t = \frac{1}{2}\log_q\left(\frac{B^2}{R}\right)$. We begin at the iterate $x_0$, where the coordinate $(x_0)_i = q^i$ for all $i < t$, and $(x_0)_i = 0$ for $i \geq t$. Observe that $\|x_0 - x^\star\|^2 \leq \sum_{i = t}^{\infty} q^{2i} \leq \frac{q^{2t}}{1 - q^2} \leq R q^{2t} \leq B^2$.
    
    Observe that for any zero-respecting algorithm, on odd machines, if for any $i$, we have $x_{2i}^m = x_{2i + 1}^m = 0$, then after any number of local computations, we still have $x_{2i + 1} = 0$. Similarly, on even machines, if for any $i$, we have $x_{2i - 1}^m = x_{2i}^m = 0$, then after any number of local computations, we still have $x_{2i} = 0$. 

    Thus, after $R$ rounds of communication, on all machines, we have $x_i^m = 0$, for all $i > t + R$. Thus for $d$ sufficiently large, we have $\|\hat{x} - x^\star\|^2 \geq \sum_{i = t +  R + 1}^d{q^{2i}} \geq \frac{q^{2t + 2R + 2} - q^{2d}}{1 - q^2} = \Omega\left(B^2 q^{2R + 2}\right) = \Omega(B^2)$ since $q = 1 - \frac{1}{R}$.

    Now observe that the Hessian of $F$ is a tridiagonal Toeplitz matrix with diagonal entries $H (q^2 + 1)$ and off-diagonal entries $-Hq$. It is well-known (see e.g., \cite{golub2005cme}) that the $d$ eigenvalues of $\tilde{M}$ are 
    $(1 + q^2)H + 2qH\cos\left(\frac{i \pi}{d + 1}\right)$ for $i = 1, \ldots, d$. Thus since $\cos(x) \geq -1$, we know that $F$ has strong-convexity parameter at least $H(q^2 + 1 - 2q) = \Omega\left(\frac{H}{R^2}\right) $, so we have $F(\hat{x}) - F(x^*) \geq \Omega\left(B^2\right)\Omega\left(\frac{H}{R^2}\right),$ which gives the desired result.
\end{proof}

\section{On the Assumption of \cite{wang2022unreasonable}}\label{app:wang}
We will now consider the first-order assumption introduced by \citet{wang2022unreasonable}. As discussed in Section \ref{sec:setting}, Assumption \ref{ass:zeta_everywhere} is very restrictive when $\xxx = \rr^d$ . \citet{wang2022unreasonable} claim that even with $\xxx =  S^\star$, the optima of $F(x)$, Assumption \eqref{ass:zeta_everywhere} (and by implication Assumption \ref{ass:relaxed_zeta_everywhere}) can be restrictive in some settings. They propose an alternative assumption that instead tries to capture how much the local iterates move when initialized at an optimum of the averaged function $F$. 
\begin{assumption}[Movement at Optima]\label{ass:rho}
Given an inner step-size $\eta$, local steps $K$, and $x^\star\in \arg\min_{x\in \rr^d}F(x)$ assume, 
    $$\frac{1}{M\eta K}\norm{\sum_{m\in[M]} x^\star - \hat{x}^m_K} \leq \rho,$$
where $\hat{x}^m_K$ is the iterate on machine $m$ after making $K$ local steps (using exact gradients) initialized at $x^\star$.
\end{assumption}
Unlike all other assumptions, note that $\rho$ in Assumption \ref{ass:rho} can be a function of the hyper-parameters $\eta$ and $K$ despite normalizing with $\eta K$. \citet{wang2022unreasonable} argue that Assumption \ref{ass:rho} is much less restrictive than the other first-order assumptions (Assumption \ref{ass:zeta_everywhere} with $\xxx=\rr^d$ or with $\xxx=S^\star$). And, when the client's objectives are strongly convex, they show a provable domination over large-mini-batch SGD in a regime of low heterogeneity (see the discussion in Section \ref{sec:lower}). However, it is unclear if the assumption is useful in the general convex setting, where we often empirically see that local SGD outperforms mini-batch SGD. We show in Section \ref{sec:lower} that this assumption is not useful in the convex setting because of the following connection with Assumption \ref{ass:zeta_everywhere} for $\xxx=S^\star$.
\begin{proposition}\label{prop:rho_and_zeta_star}
If the functions of the machines $\{F_m\}_{m\in[M]}$ satisfy Assumption \ref{ass:zeta_everywhere} for $\xxx=S^\star$ with $\zeta_\star$, then we have,
$$\norm{\frac{1}{M\eta K}\sum_{m\in[M]} x^\star - \hat{x}^m_K} \leq \zeta_\star\rb{\rb{1+\eta H}^{K-1}-1}.$$ 
\end{proposition}
\begin{proof}
Define $\phi(k) := \norm{\frac{1}{M\eta K}\sum_{m\in[M]} x^\star - \hat{x}^m_k}$ where $\hat{x}^m_k$ is the $k$th gradient descent iterate on machine $m$ initialized at $\hat{x}^m_0 = x^\star$. Then note the following,
\begin{align}
    \phi(K) &= \norm{\frac{1}{M\eta K}\sum_{m\in[M]} x^\star - \hat{x}^m_K},\nonumber\\
    &= \norm{\frac{1}{M\eta K}\sum_{m\in[M]} x^\star - \hat{x}^m_{K-1} + \eta\nabla F_m(\hat{x}^m_{K-1})},\nonumber\\
    &\leq \norm{\frac{1}{M\eta K}\sum_{m\in[M]} x^\star - \hat{x}^m_{K-1}} + \norm{\frac{1}{MK}\sum_{m\in[M]}\nabla F_m(\hat{x}^m_{K-1})},\nonumber\\
    &= \phi(K-1) + \norm{\frac{1}{MK}\sum_{m\in[M]}\nabla F_m(\hat{x}^m_{K-1}) - \nabla F_m(x^\star)},\nonumber\\
    &\leq \phi(K-1) + \frac{1}{MK}\sum_{m\in[M]}\norm{\nabla F_m(\hat{x}^m_{K-1}) - \nabla F_m(x^\star)},\nonumber\\
    &\leq \phi(K-1) + \frac{H}{MK}\sum_{m\in[M]}\norm{\hat{x}^m_{K-1} - x^\star},\nonumber\\
    &= \phi(K-1) + \frac{H}{K}\delta(K-1), \label{eq:recursion_1}
\end{align}
where we define $\delta(k) := \frac{1}{M}\sum_{m\in[M]}\norm{\hat{x}^m_{k} - x^\star}$. Now we consider another recursion on $\delta(k)$ to introduce the $\zeta_\star$ assumption:
\begin{align}
    \delta(k) &= \frac{1}{M}\sum_{m\in[M]}\norm{\hat{x}^m_{k} - x^\star},\nonumber\\
    &\leq \frac{1}{M}\sum_{m\in[M]}\norm{\hat{x}^m_{k-1} - x^\star} + \frac{\eta}{M}\sum_{m\in[M]}\norm{\nabla F_m(\hat{x}_m^{k-1})},\nonumber\\
    &\leq \frac{1}{M}\sum_{m\in[M]}\norm{\hat{x}^m_{k-1} - x^\star} + \frac{\eta}{M}\sum_{m\in[M]}\rb{\norm{\nabla F_m(\hat{x}_m^{k-1}) - \nabla F_m(x^\star)} + \norm{\nabla F_m(x^\star)}},\nonumber\\
    &\leq \frac{(1+\eta H)}{M}\sum_{m\in[M]}\norm{\hat{x}^m_{k-1} - x^\star} + \eta \zeta_\star,\nonumber\\
    &\leq (1+\eta H)\delta(k-1) + \eta \zeta_\star,\nonumber\\
    &\leq \frac{\zeta_\star}{H}\rb{\rb{1+\eta H}^k-1}. \qquad \qquad (\delta(0)=0)\label{eq:recursion_2}
\end{align}
Plugging \eqref{eq:recursion_2} back into \ref{eq:recursion_1} we get,
\begin{align*}
    \phi(K) &\leq \phi(K-1) + \frac{\zeta_\star}{K}\rb{\rb{1+\eta H}^{K-1}-1},\\
    &= \zeta_\star\rb{\rb{1+\eta H}^{K-1}-1},
\end{align*}
which proves the claim.
\end{proof}
\begin{remark}
    When Assumption \ref{ass:zeta_everywhere} for $\xxx=S^\star$ can be satisfied with $\zeta_\star=0$, we can satisfy Assumption \ref{ass:rho} with $\rho=0$.    
\end{remark}

As might be evident already, Proposition \ref{prop:lb_zeta0} has implications for the heterogeneity assumption by \cite{wang2022unreasonable}. This is because the construction used in the proposition satisfies Assumption \ref{ass:rho} with $\rho=0$. In particular, in the general convex setting under Assumption \ref{ass:rho}, local SGD can not hope to benefit in the optimization term from a large $K$. This implies that it is \textbf{not possible} to explain the effectiveness of local SGD by controlling the movement of the local iterates at the optima, as postulated by \cite{wang2022unreasonable}. Note that Theorem \ref{thm:AIlb_zeta0} also applies to Assumption \ref{ass:rho}. Thus, large mini-batch SGD is also min-max optimal under Assumptions \ref{ass:smth}, \ref{ass:bounded_optima}, and \ref{ass:rho} for convex objectives.

To reconcile this negative result with the upper bound of \cite{wang2022unreasonable}, we note that their analyses crucially rely on the strong convexity of each machine's objective and not just the strong convexity of the average function, something which is lacking in our hard instance in Proposition \ref{prop:lb_zeta0}. This highlights that controlling the client drift for local SGD between communication rounds in the general convex setting is much more challenging due to potential directions of no information on each machine (both $A_1, A_2$ are rank one in Proposition \ref{prop:lb_zeta0}).

\subsection{Beating Mini-batch SGD}

Finally, we note that for quadratic functions of the form discussed in Section \ref{sec:fixed}, we can bound the average movement at optimum as follows,

\begin{align*}
    \rho &= \frac{1}{M\eta K}\norm{\sum_{m\in[M]} x^\star - x_{K}^m} = \frac{1}{M\eta K}\norm{\sum_{m\in[M]} x^\star - x_m^\star - (I-\eta A_m)^K(x_\star-x_m^\star)}\enspace,\\
    &= \frac{1}{M\eta K}\norm{\sum_{m\in[M]} \rb{I- (I-\eta A_m)^K}(x_\star-x_m^\star)}\enspace,\\
    &\leq \frac{1}{M\eta K}\sum_{m\in[M]} \norm{ \rb{I- (I-\eta A_m)^K}}\norm{x_\star-x_m^\star}\leq \frac{1-(1-\eta L)^K}{\eta K}\zeta_\star\enspace.
\end{align*}
If $\eta = \frac{1}{2HK^p}$ for $p<1$, and $K\to \infty$ we get that,
\begin{align*}
    \lim_{K\to \infty} \rho &\leq \lim_{K\to \infty} \frac{1-(1-\eta H)^K}{\eta K}\zeta_\star\enspace,\\
    &\leq \lim_{K\to \infty} \frac{1-\rb{1-\frac{1}{2K^p}}^K}{K^{1-p}}\cdot 2L\zeta_\star\enspace,\\
    &= 0\enspace.
\end{align*}
On the other hand, if $\eta=\frac{1}{2HK}$, then we get that,
\begin{align*}
    \lim_{K\to \infty} \rho &\leq \lim_{K\to \infty} \frac{1-(1-\eta H)^K}{\eta K}\zeta_\star\enspace,\\
    &\leq \lim_{K\to \infty} \rb{1-\rb{1-\frac{1}{2K}}^K}\cdot 2L\zeta_\star\enspace,\\
    &= \rb{1-\frac{1}{\sqrt{e}}}\cdot 2L\zeta_\star\enspace.
\end{align*}

This suggests that the Corollary 2 in \cite{wang2022unreasonable} must use $\eta = \Omega\rb{\frac{1}{K}}$ to ensure that $\rho$ can reach arbitrarily small values. Unfortunately the corollary can only be obtained by setting $\eta < \min\cb{\frac{1}{\mu K}, \frac{1}{L}}$. Thus, in their existing result, it seems \textbf{impossible} to improve the convergence w.r.t. MB-SGD unconditionally. This also highlights that our result in Theorem \ref{thm:fixed_convergence} is much cleaner, depends on only problem-dependentth parameters, and can improve over mini-batch SGD. 

\section{Missing Proofs from Section \ref{sec:upper}}\label{app:upper proof}
\subsection{Proof of Poposition \ref{prop:zeta_bound}}
\begin{proof}
    Denote the function $G_m := F_m - F$ for all $m\in [M]$. Then note that using Taylor expansion, we can write for any $x\in\bb_2(D)$ and $x^\star\in S$,
    \begin{align*}
        \nabla G_m(x) - \nabla G_m(x^\star) = \sb{\int_{0}^1\nabla^2 G_m(x^\star + s(x-x^\star))ds}(x-x^\star)\enspace.
    \end{align*}
    Re-arranging, taking norm, and applying triangle inequality implies,
    \begin{align*}
        \norm{\nabla G_m(x)}  &= \norm{\nabla G_m(x^\star)  + \sb{\int_{0}^1\nabla^2 G_m(x^\star + s(x-x^\star))ds}(x-x^\star)}\enspace,\\
        &\leq \norm{\nabla F_m(x^\star)} + \norm{\int_{0}^1\nabla^2 G_m(x^\star + s(x-x^\star))ds}\norm{x-x^\star}\enspace,\\
        &\leq \norm{\nabla F_m(x^\star)} + \sb{\int_{0}^1\norm{\nabla^2 G_m(x^\star + s(x-x^\star))}ds}\norm{x-x^\star}\enspace,\\
        &\leq \norm{\nabla F_m(x^\star)} + \sb{\int_{0}^1\tau ds}D\enspace,\\
        &\leq \norm{\nabla F_m(x^\star)} + \tau D\enspace.
    \end{align*}
    Averaging over the machines and noting the definition of $\zeta_\star$ gives us the desired result.
\end{proof}

\subsection{Convergence in the Strongly Convex Setting}
Recall that the average objective $F$ is $\mu$-strongly convex, if for all $x,y\in \rr^d$, 
$$F(y) + \inner{\nabla F(y)}{x-y} + \frac{\mu}{2}\norm{x-y}^2 \leq F(x).$$
We can give the following upper bound for local SGD for optimizing problems in the class $\ppp^{H,Q,B,\sigma}_\tau$ with $\mu$-strongly convex average objective.

\begin{theorem}[Informal, Strongly Convex]\label{thm: LSGD_upper_bound}
    For any $K,R, M \geq 1$ and $ H, B , Q, \sigma, \tau,\geq 0,\  \mu > 0$ consider a problem instance in the class $\ppp^{H,Q,B,\sigma}_\tau$, with $\mu$-strongly convex average objective $F$. Then there exists a sequence of weights $w_{r,k}\geq 0$, for $r\in[R], k\in[0,K-1]$ and a fixed inner stepsize of $\eta \leq \frac{1}{2H}$ such that with outer stepsize of $\beta = 1$ we have the following convergence rate (up to logarithmic factors) for local SGD where $W_{R,K} = \sum_{r\in[R], k\in[0,K-1]}w_{r,k}$: 
\begin{align*}
     \mathbb{E} &\left[F\left(\frac{1}{MW_{R,K}} \sum_{m=1}^M \sum_{r=1}^{R}\sum_{k=0}^{K-1} w_{r,k} x^m_{r,k}\right) - F(x^{\star}) + \mu \norm{\frac{1}{M}\sum_{m\in[M]} x_{R,K}^m - x^\star}^2\right] \preceq\\ 
     &\qquad \qquad \qquad \qquad HB^2 \exp \left(-\frac{\mu KR}{H}\right) + \frac{\sigma^2}{\mu M KR}+ \frac{\tau^2\sigma^2}{\mu^3 KR^2} + \frac{\tau^2 \zeta^2}{\mu^3 R^2}+   \frac{Q^2\sigma^4}{\mu^5 K^2R^4} + \frac{Q^2\zeta^4}{\mu^5 R^4} \,.
\end{align*}
Proof of this Theorem can be found in \ref{app:strongly convex upper bound proof}.
\end{theorem}

\subsection{Useful Lemmas}\label{app:lemmas}

In this section, we provide a list of useful lemmas which will be used in the proofs. We use the notation $\mathbb{E}_t$ for expectation conditioned on $x^1_t,...,x^M_t$ and $\mathbb{E}$ for the unconditional expectation. 

\begin{lemma} \label{Lemma: jensen's inequality}
For a convex function $F$ we have:
\begin{equation}
    F\left(\frac 1 M \sum_{m=1} ^{M}x_m\right) \leq \frac 1 M \sum_{m=1} ^{M} F(x_m)  \,.
\end{equation}
\end{lemma}

\begin{lemma}\label{Lemma: base triangle inequality}
For a set of $M$ vectors $a_1, a_2, ..., a_M \in \mathbb{R}^d$ we have:
\begin{equation}
    \norm{\sum_{m=1} ^{M} a_m} \leq \sum_{m=1} ^{M} \norm{a_m}  \,.
\end{equation}
\end{lemma}

\begin{lemma}\label{Lemma: triangle inequality}
For a set of $M$ vectors $a_1, a_2, ..., a_M \in \mathbb{R}^d$ we have:
\begin{equation}
    \norm{\sum_{m=1} ^{M} a_m}^2 \leq M \sum_{m=1} ^{M} \norm{a_m}^2  \,.
\end{equation}
\end{lemma}

\begin{lemma} \label{Lemma: generalized triangle inequality}
    For two arbitrary vectors $a,b \in \mathbb{R}^d$ and $\forall \gamma > 0$ we have:
    \begin{equation}
        \norm{a+b}^2 \leq (1+ \gamma ) \norm{a}^2 + (1+ \gamma^{-1} ) \norm{b}^2 \,.
    \end{equation} 
\end{lemma}

\begin{lemma} \label{lemma: noise of gradient differences}
    Let Assumption \ref{ass:stoch_first_order} hold. Then we have:
    \begin{align}
        \mathbb{E}_t \norm{\frac 1 M \sum_{m=1} ^{M} g^m_t - \frac 1 M \sum_{m=1} ^{M} \nabla F_m(x^m_t)}^2 \leq \frac{\sigma^2}{M} \,.
    \end{align}
\end{lemma} 

\begin{lemma} \label{lemma: co-coercivity of gradients}
    Let $F$ be a convex and $H$-smooth function. Then for any $x,y \in \mathbb{R}^d$ we have:
    \begin{equation}
        \frac{1}{2H} \norm{\nabla F(x) - \nabla F(y)}^2 \leq F(y) - F(x) + \langle \nabla F(x), x-y \rangle \,.
    \end{equation}
\end{lemma}

\begin{lemma} \label{lemma: mime's inequality}
    Let Assumption \ref{ass:tau} hold and $F(x) = \frac 1 M \sum_{m=1}^M F_m(x)$. Then for any $x,y \in \mathbb{R}^d$ we have the following inequality:
    \begin{align}
        \norm{\nabla F_m(x) - \nabla F(x) + \nabla F(y) - \nabla F_m(y)}^2 \leq \tau^2 \norm{x - y}^2 \,.
    \end{align}
\end{lemma}
\begin{proof}
    We define the function $\Psi(z) =  F_m(z) -  F(z)$ so $\nabla^2 \Psi(z) = \nabla^2 F_m(z) - \nabla^2 F(z)$. By Assumption \ref{ass:tau} we know that $\forall z \in \mathbb{R}^d, \norm{\nabla^2 \Psi(z)} \leq \tau$ which means $\Psi(z)$ is $\tau$-smooth. From the smoothness property we know: 
    \begin{align*}
        \norm{\nabla \Psi(x) - \nabla \Psi(y)} &\leq \tau \norm{x-y} \,.
    \end{align*}
    By replacing the definition of $\Psi$ we get: 
    \begin{align*}
        \norm{\nabla F_m(x) - \nabla F(x) - \nabla F_m(y) + \nabla F(y)} \leq \tau \norm{x-y}  \,.
    \end{align*} 
\end{proof}

\begin{lemma}[{\citealt[Definition 2]{kovalev2019stochastic}}]\label{lemma: quadratic function approximation inequality}
    Let function $F$ satisfy Assumption \ref{ass: lipschitz hessian}, then for any $x,y \in \mathbb{R}^d$, we have the following inequality:
    \begin{align} 
        \norm{\nabla F(x) - \nabla F(y) - \nabla^2 F(y)(x-y)} \leq \frac Q 2 \norm{x - y}^2  \,.
    \end{align}
\end{lemma}

\begin{lemma}[\citet{woodworth2020minibatch} Lemma 8] \label{lemma: weight variance upper bound}
    Let Assumptions \ref{ass:stoch_first_order} and \ref{ass:zeta_everywhere} hold. We can upper bound the consensus error $\varXi_t$ using a learning rate of $\eta_t = \eta \leq \frac{1}{2H}$ as follows:
    \begin{align}
        \varXi_t := \frac 1 M \sum_{m=1}^M \ee\sb{\norm{x^m_t - \bar x_t}^2} \leq 3 K \sigma^2 \eta^2 + 6 K^2 \eta^2 \zeta^2 \,,
    \end{align}
    where $x^m_t$ is the parameter of client $m$ at time step $t$ and $\bar x_t = \frac 1 M \sum_{m=1}^M x^m_t$.
\end{lemma}
In addition to the above lemma, we would also need the following lemma that controls the fourth moment of the consensus error. This lemma is useful when proving the upper bound with third-order smoothness. 
\begin{lemma}[Consensus Error Fourth Moment]\label{lem:cons_error_fourth}
    Let Assumptions \ref{ass:stoch_first_order}, \ref{ass:stoch_first_order_fourth} and \ref{ass:zeta_everywhere} hold. Then for all $t\in[T]$ we can bound the fourth moment of the consensus error as,
    \begin{align*}
        \frac{1}{M}\sum_{m\in[M]}\ee\sb{\norm{x_t-x_t^m}^4} \leq 3840\eta^4K^4\zeta^4 + 5920\eta^4 K^2\sigma^4\enspace,
    \end{align*}
    where we just need to ensure that $\eta\leq \frac{1}{2H}$.
\end{lemma}
\begin{proof}
    First by Jensen's inequality we know:
    \begin{align*}
        \frac{1}{M}\sum_{m\in[M]}\ee\sb{\norm{x_t-x_t^m}^4} \leq \frac{1}{M^2}\sum_{n\in[M]}\sum_{m\in[M]}\ee\sb{\norm{x_t^n-x_t^m}^4}
    \end{align*}
    Note the following about the fourth moment of the difference between the iterates on two machines $m,n\in[M]$ (where we denote $g_t^m := \nabla f(x_t^m; z_t^m)$ is the $t$-th stochastic gradient on machine $m$ and $\xi_{t}^m:=\nabla F_m(x_t^m) - g_t^m$),
    \begin{align*}
        &\ee\sb{\norm{x_t^m - x_t^n}^4}\\
        &= \ee\sb{\norm{x_{t-1}^m - x_{t-1}^n - \eta g_{t-1}^m +\eta g_{t-1}^n}^4}\enspace,\\
        &= \ee\sb{\rb{\norm{x_{t-1}^m - x_{t-1}^n - \eta \nabla F_m(x_{t-1}^m) + \eta \nabla F_n(x_{t-1}^n) + \eta \xi_{t-1}^m - \eta \xi_{t-1}^n}^2}^2}\enspace,\\
        &= \ee\Bigg[\bigg(\norm{x_{t-1}^m - x_{t-1}^n - \eta \nabla F_m(x_{t-1}^m) + \eta \nabla F_n(x_{t-1}^n)}^2 + \eta^2 \norm{\xi_{t-1}^m -\xi_{t-1}^n}^2\\
        &\quad + 2\eta\inner{x_{t-1}^m - x_{t-1}^n - \eta \nabla F_m(x_{t-1}^m) + \eta \nabla F_n(x_{t-1}^n)}{\xi_{t-1}^m - \xi_{t-1}^n}\bigg)^2\Bigg]\enspace,\\
        &= \ee\sb{\norm{x_{t-1}^m - x_{t-1}^n - \eta \nabla F_m(x_{t-1}^m) + \eta \nabla F_n(x_{t-1}^n)}^4} + \eta^4 \ee\sb{\norm{\xi_{t-1}^m -\xi_{t-1}^n}^4}\\
        &\quad +4\eta^2\ee\sb{\rb{\inner{x_{t-1}^m - x_{t-1}^n - \eta \nabla F_m(x_{t-1}^m) + \eta \nabla F_n(x_{t-1}^n)}{\xi_{t-1}^m - \xi_{t-1}^n}}^2}\\
        &\quad + 2\eta^2\ee\sb{\norm{x_{t-1}^m - x_{t-1}^n - \eta \nabla F_m(x_{t-1}^m) + \eta \nabla F_n(x_{t-1}^n)}^2\norm{\xi_{t-1}^m -\xi_{t-1}^n}^2}\\
        &\quad +4\eta^3\ee\sb{\inner{x_{t-1}^m - x_{t-1}^n - \eta \nabla F_m(x_{t-1}^m) + \eta \nabla F_n(x_{t-1}^n)}{\xi_{t-1}^m - \xi_{t-1}^n}\norm{\xi_{t-1}^m - \xi_{t-1}^n}^2}\\
        &\quad + 4\eta\ee\sb{\norm{x_{t-1}^m - x_{t-1}^n - \eta \nabla F_m(x_{t-1}^m) + \eta \nabla F_n(x_{t-1}^n)}^2\rb{x_{t-1}^m - x_{t-1}^n - \eta \nabla F_m(x_{t-1}^m) + \eta \nabla F_n(x_{t-1}^n)}}^T\cancelto{0}{\ee\sb{\xi_{t-1}^m - \xi_{t-1}^n}}\enspace,\\
        &\leq^{\text{(C.S. Inequality, Assumption in \eqref{ass:stoch_first_order_fourth})}} \ee\sb{\norm{x_{t-1}^m - x_{t-1}^n - \eta \nabla F_m(x_{t-1}^m) + \eta \nabla F_n(x_{t-1}^n)}^4} + 8\sigma^4\eta^4\\
        &\quad + 6\eta^2\ee\sb{\norm{x_{t-1}^m - x_{t-1}^n - \eta \nabla F_m(x_{t-1}^m) + \eta \nabla F_n(x_{t-1}^n)}^2}\ee\sb{\norm{\xi_{t-1}^m -\xi_{t-1}^n}^2}\\
        &\quad +4\eta^3\ee\sb{\norm{x_{t-1}^m - x_{t-1}^n - \eta \nabla F_m(x_{t-1}^m) + \eta \nabla F_n(x_{t-1}^n)}}\ee\sb{\norm{\xi_{t-1}^m - \xi_{t-1}^n}^3}\enspace,
\end{align*}
In order to bound the term $\ee \sb{\norm{\xi^m_{t-1} - \xi^n_{t-1}}^3}$ we use Cauchy-Schwarz Inequality:
\begin{align*}
    \ee \sb{\norm{\xi^m_{t-1} - \xi^n_{t-1}}^3} &= \ee \sb{\norm{\xi^m_{t-1} - \xi^n_{t-1}}\cdot \norm{\xi^m_{t-1} - \xi^n_{t-1}}^2} \\
    &\leq \sqrt{\ee \sb{\norm{\xi^m_{t-1} - \xi^n_{t-1}}^2} \ee \sb{\norm{\xi^m_{t-1} - \xi^n_{t-1}}^4}} \stackrel{\text{Assumptions (\ref{ass:stoch_first_order}, \ref{ass:stoch_first_order_fourth})}}{\leq} 4\sqrt{\sigma^6}
\end{align*}
Also the term $\ee\sb{\norm{x_{t-1}^m - x_{t-1}^n - \eta \nabla F_m(x_{t-1}^m) + \eta \nabla F_n(x_{t-1}^n)}}$ can be bounded as: 
\begin{align*}
    \ee\sb{\norm{x_{t-1}^m - x_{t-1}^n - \eta \nabla F_m(x_{t-1}^m) + \eta \nabla F_n(x_{t-1}^n)}} \stackrel{\text{Jensen's Inequality}}{\leq} \sqrt{\ee\sb{\norm{x_{t-1}^m - x_{t-1}^n - \eta \nabla F_m(x_{t-1}^m) + \eta \nabla F_n(x_{t-1}^n)}^2}}
\end{align*}
Putting everything back together gives us: 
\begin{align*}
        &\ee \sb{\norm{x^n_t - x^m_t}^4} \\
        &\stackrel{\text{(Assumption in \eqref{ass:stoch_first_order})}}{\leq} \ee\sb{\norm{x_{t-1}^m - x_{t-1}^n - \eta \nabla F_m(x_{t-1}^m) + \eta \nabla F_n(x_{t-1}^n)}^4} + 8\eta^4 \sigma^4\\
        &\quad +12\eta^2\sigma^2\ee\sb{\norm{x_{t-1}^m - x_{t-1}^n - \eta \nabla F_m(x_{t-1}^m) + \eta \nabla F_n(x_{t-1}^n)}^2}\\
        &\quad +16\eta^3\sqrt{\sigma^6 \ee\sb{\norm{x_{t-1}^m - x_{t-1}^n - \eta \nabla F_m(x_{t-1}^m) + \eta \nabla F_n(x_{t-1}^n)}^2}}\enspace,
\end{align*}
To bound the third term in the above inequality, we use the A.M. - G.M. Inequality $\sqrt{ab} \leq \frac{a}{2\gamma} + \frac{\gamma b}{2}$ for $\gamma >0$. Let $\gamma = \eta, a=\sigma^2 \ee\sb{\norm{x_{t-1}^m - x_{t-1}^n - \eta \nabla F_m(x_{t-1}^m) + \eta \nabla F_n(x_{t-1}^n)}^2}, b=\sigma^4$. We have:
\begin{align*}
    &16\eta^3\sqrt{\sigma^6 \ee\sb{\norm{x_{t-1}^m - x_{t-1}^n - \eta \nabla F_m(x_{t-1}^m) + \eta \nabla F_n(x_{t-1}^n)}^2}}\\
    &= 16\eta^3\sqrt{(\sigma^4) \left(\sigma^2 \ee\sb{\norm{x_{t-1}^m - x_{t-1}^n - \eta \nabla F_m(x_{t-1}^m) + \eta \nabla F_n(x_{t-1}^n)}^2}\right)} \\
    &\leq 16\eta^3 \left(\frac{\eta \sigma^4}{2} + \frac{\sigma^2}{2\eta}\ee\sb{\norm{x_{t-1}^m - x_{t-1}^n - \eta \nabla F_m(x_{t-1}^m) + \eta \nabla F_n(x_{t-1}^n)}^2} \right)
\end{align*}
So we have: 
\begin{align*}
        &\ee \sb{\norm{x^n_t - x^m_t}^4} \\
        &\leq \ee\sb{\norm{x_{t-1}^m - x_{t-1}^n - \eta \nabla F_m(x_{t-1}^m) + \eta \nabla F_n(x_{t-1}^n)}^4} + 8\eta^4 \sigma^4 +12\eta^2\sigma^2\ee\sb{\norm{x_{t-1}^m - x_{t-1}^n - \eta \nabla F_m(x_{t-1}^m) + \eta \nabla F_n(x_{t-1}^n)}^2}\\
        &\quad +16\eta^3\rb{\frac{\eta\sigma^4}{2} +  \frac{\sigma^2}{2\eta}\ee\sb{\norm{x_{t-1}^m - x_{t-1}^n - \eta \nabla F_m(x_{t-1}^m) + \eta \nabla F_n(x_{t-1}^n)}^2}}\enspace,\\
        &= \ee\sb{\norm{x_{t-1}^m - x_{t-1}^n - \eta \nabla F_m(x_{t-1}^m) + \eta \nabla F_n(x_{t-1}^n)}^4} +20\eta^2\sigma^2\ee\sb{\norm{x_{t-1}^m - x_{t-1}^n - \eta \nabla F_m(x_{t-1}^m) + \eta \nabla F_n(x_{t-1}^n)}^2}+ 16\eta^4 \sigma^4\\
        &= \ee\sb{\norm{x_{t-1}^m - x_{t-1}^n - \eta \nabla F(x_{t-1}^m) + \eta\nabla F(x_{t-1}^n) + \eta \rb{\nabla F(x_{t-1}^m) - \nabla F_m(x_{t-1}^m)} - \eta \rb{\nabla F(x_{t-1}^n) - \nabla F_n(x_{t-1}^n)}}^4}\\
        &\quad +20\eta^2\sigma^2\ee\sb{\norm{x_{t-1}^m - x_{t-1}^n - \eta \nabla F(x_{t-1}^m) + \eta\nabla F(x_{t-1}^n) + \eta \rb{\nabla F(x_{t-1}^m) - \nabla F_m(x_{t-1}^m)} - \eta \rb{\nabla F(x_{t-1}^n) - \nabla F_n(x_{t-1}^n)}}^2}\\
        &\quad + 16\eta^4 \sigma^4
\end{align*}
Now by using Lemma \ref{Lemma: generalized triangle inequality} and generalized triangle inequality for fourth moment $\norm{a+b}^4 \leq \left(1+\frac{1}{\gamma} \right)^3\norm{a}^4 + (1+\gamma)^3 \norm{b}^4$ by choosing $\gamma = K-1$ we have: 
\begin{align*}
        &\leq \rb{1+\frac{1}{K-1}}^3\ee\sb{\norm{x_{t-1}^m - x_{t-1}^n - \eta \nabla F(x_{t-1}^m) + \eta\nabla F(x_{t-1}^n)}^4}\\ 
        &\quad + 2\eta^4K^3\ee\sb{\norm{\rb{\nabla F(x_{t-1}^m) - \nabla F_m(x_{t-1}^m)} - \rb{\nabla F(x_{t-1}^n) - \nabla F_n(x_{t-1}^n)}}^4}\\
        &\quad + 20\eta^2\sigma^2\rb{1+ \frac{1}{K-1}}\ee\sb{\norm{x_{t-1}^m - x_{t-1}^n - \eta \nabla F(x_{t-1}^m) + \eta\nabla F(x_{t-1}^n)}^2}\\
        &\quad + 20\eta^4\sigma^2K\ee\sb{\norm{\rb{\nabla F(x_{t-1}^m) - \nabla F_m(x_{t-1}^m)} - \rb{\nabla F(x_{t-1}^n) - \nabla F_n(x_{t-1}^n)}}^2}+ 16\eta^4 \sigma^4
\end{align*}
From the mean-value theorem we know that $\nabla F(x) - \nabla F(y) = \nabla^2 F(c)(x-y)$ for some $c = \lambda x + (1-\lambda) y$. By applying this theorem to the first and third term of the above inequality we have: 
\begin{align*}
    &\rb{1+\frac{1}{K-1}}^3\ee\sb{\norm{x_{t-1}^m - x_{t-1}^n - (\eta \nabla F(x_{t-1}^m) - \eta\nabla F(x_{t-1}^n))}^4} \\
    &= \rb{1+\frac{1}{K-1}}^3\ee\sb{\norm{x_{t-1}^m - x_{t-1}^n - \eta \nabla^2 F(c)(x_{t-1}^m-x_{t-1}^n)}^4} \\
    &= \rb{1+\frac{1}{K-1}}^3\ee\sb{\norm{(I- \eta \nabla^2 F(c))(x_{t-1}^m - x_{t-1}^n)}^4} \\
    &\leq \rb{1+\frac{1}{K-1}}^3 (1- \eta \mu)^4 \ee \sb{\norm{x_{t-1}^m-x_{t-1}^n}^4}
\end{align*}
With the same approach for the third term we have: 
\begin{align*}
    &20\eta^2\sigma^2\rb{1+ \frac{1}{K-1}}\ee\sb{\norm{x_{t-1}^m - x_{t-1}^n - \eta \nabla F(x_{t-1}^m) + \eta\nabla F(x_{t-1}^n)}^2} \\
    &\leq 20\eta^2\sigma^2\rb{1+ \frac{1}{K-1}}(1-\eta \mu)^2 \ee\sb{\norm{x_{t-1}^m - x_{t-1}^n}^2}
\end{align*}
Putting all together gives us: 
\begin{align*}
        &\ee \sb{\norm{x^n_t - x^m_t}^4} \\
        &\leq \rb{1+\frac{1}{K-1}}^3(1-\eta\mu)^4\ee\sb{\norm{x_{t-1}^m - x_{t-1}^n}^4}\\ 
        &\quad + 16\eta^4K^3\rb{\ee\sb{\norm{\nabla F(x_{t-1}^m) - \nabla F_m(x_{t-1}^m)}^4} + \ee\sb{\norm{\nabla F(x_{t-1}^n) - \nabla F_n(x_{t-1}^n)}^4}}\\
        &\quad + 20\eta^2\sigma^2\rb{1+ \frac{1}{K-1}}(1-\eta\mu)^2\ee\sb{\norm{x_{t-1}^m - x_{t-1}^n}^2}\\
        &\quad + 40\eta^4\sigma^2K\rb{\ee\sb{\norm{\nabla F(x_{t-1}^m) - \nabla F_m(x_{t-1}^m)}^2} + \ee\sb{\norm{\nabla F(x_{t-1}^n) - \nabla F_n(x_{t-1}^n)}^2}}+ 16\eta^4 \sigma^4\enspace,\\
        &\leq \rb{1+\frac{1}{K-1}}^3(1-\eta\mu)^4\ee\sb{\norm{x_{t-1}^m - x_{t-1}^n}^4} + 32\eta^4K^3\zeta^4\\
        &\quad + 20\eta^2\sigma^2\rb{1+ \frac{1}{K-1}}(1-\eta\mu)^2\ee\sb{\norm{x_{t-1}^m - x_{t-1}^n}^2} + 80\eta^4\sigma^2K\zeta^2+ 16\eta^4 \sigma^4\enspace,\\
        &\stackrel{\text{(Lemma \ref{lemma: weight variance upper bound})}}{\leq} \rb{1+\frac{1}{K-1}}^3\ee\sb{\norm{x_{t-1}^m - x_{t-1}^n}^4} + 32\eta^4K^3\zeta^4\\
        &\quad + 40\eta^2\sigma^2\rb{3K\sigma^2\eta^2 + 6K^2\eta^2\zeta^2} + 80\eta^4\sigma^2K\zeta^2+ 16\eta^4 \sigma^4\enspace,\\
        &\leq \rb{1+\frac{1}{K-1}}^3\ee\sb{\norm{x_{t-1}^m - x_{t-1}^n}^4} + 32\eta^4K^3\zeta^4 + 136\eta^4 K\sigma^4 +320 \eta^4\sigma^2K^2\zeta^2\enspace,\\
        &\leq \rb{1+\frac{1}{K-1}}^{3(K-1)}\rb{32\eta^4K^4\zeta^4 + 136\eta^4 K^2\sigma^4 +320 \eta^4\sigma^2K^3\zeta^2}\enspace,\\
        &\leq^{\text{(a)}} 20\rb{32\eta^4K^4\zeta^4 + 136\eta^4 K^2\sigma^4 +160\cdot2\cdot\rb{\eta^2\sigma^2K}\cdot\rb{\eta^2\zeta^2K^2}}\enspace,\\
        &\leq 20\rb{192\eta^4K^4\zeta^4 + 296\eta^4 K^2\sigma^4}\enspace,\\
        &\leq 3840\eta^4K^4\zeta^4 + 5920\eta^4 K^2\sigma^4\enspace,
    \end{align*}
    where in (a) we used that $(1+1/x)^x \leq 20$ for all $x\geq0$. 
    Finally averaging this over $m,n\in[M]$ implies,
    \begin{align*}
        \frac{1}{M}\sum_{m\in[M]}\ee\sb{\norm{x_t-x_t^m}^4} \leq \frac{1}{M^2}\sum_{m,n\in[M]}\ee\sb{\norm{x_t^n-x_t^m}^4} \leq 3840\eta^4K^4\zeta^4 + 5920\eta^4 K^2\sigma^4\enspace,
    \end{align*}
    which proves the lemma.
\end{proof}

\begin{lemma}[Stepsize Tuning] \label{lemma: stepsize}
    Our upper bound has the following recursive form: 
    \begin{align}
        r_{t+1} \leq (1-A\eta_t)r_t - \eta_t e_t + C\eta_t^2 + E\eta_t^3 + G \eta_t^5 \,, \label{eq:1353}
    \end{align}
    where $\{r_t\}_{t\geq 0}$ and $\{e_t\}_{t\geq 0}$ are two non-negative sequences and $\eta_t \leq \frac{1}{D}$ for a $D>0$. With the following choice of parameters we recover our recursive upper bound used in the proof of Theorem \ref{thm: LSGD_upper_bound} (see below):
    \begin{align*}
        r_t = \norm{\bar x_t - x^\star}^2,\quad e_t = \Bigr[ F(\bar x_t) - F(x^\star) \Bigr],\quad C = \frac{\sigma^2}{M},\quad D=2H,\quad A = \frac \mu 2  \,, \\
        E=\frac{24 K \tau^2 \sigma^2}{\mu} + \frac{48 K^2 \tau^2 \zeta^2}{\mu}\,, \quad G=\frac{36K^2 Q^2 \sigma^4 }{\mu} + \frac{144 Q^2 K^4 \zeta^4 }{\mu} \,.
    \end{align*}  
\end{lemma}
The recursion~\eqref{eq:1353} can be solved as follows: 
\begin{align*}
    \frac{1}{W_T}\sum_{t=0}^T w_t e_t + A r_{T+1} &\leq D r_0 \exp{\left(\frac{-A T}{ D} \right)}+ \tilde{\mathcal{O}} \left(\frac{C}{AT} \right) + \tilde{\mathcal{O}} \left(\frac{E}{A^2T^2} \right) +  \tilde{\mathcal{O}} \left(\frac{G}{A^4T^4} \right) \,,
\end{align*} 
where we define the coefficient $w_t = (1-A\eta)^{-(t+1)}$ and $W_T = \sum_{t=0}^T w_t$.

\begin{proof}
The proof is similar to the works \citep{stich2019unified,koloskova2020unified}. We divide both sides of the recursive inequality by $\eta_t$ and rearrange the terms: 
\begin{align*}
    e_t 
    &\leq \frac{(1-A\eta_t)}{\eta_t}r_t - \frac{1}{\eta_t}r_{t+1} + C\eta_t + E\eta_t^2 + G\eta_t^4
\end{align*}
 Note that learning rate is fixed $\eta_t = \eta$. Then we multiply both sides of the inequality by $w_t$ and sum over $t$ which gives us: 
\begin{align*}
    \frac{1}{W_T}\sum_{t=0}^T w_t e_t &\leq  \frac{1}{W_T}\sum_{t=0}^T  \left[\frac{(1-A\eta)^{-(t+1)}(1-A\eta)}{\eta}r_t - \frac{(1-A\eta)^{-(t+1)}}{\eta}r_{t+1}\right] + C\eta + E\eta^2 + G\eta^4 \\
    &=  \frac{1}{W_T}\sum_{t=0}^T  \left[\frac{(1-A\eta)^{-t}}{\eta}r_t - \frac{(1-A\eta)^{-(t+1)}}{\eta}r_{t+1}\right] + C\eta + E\eta^2 + G\eta^4 \\
    &=  \frac{1}{W_T}\sum_{t=0}^T  \left[\frac{w_{t-1}}{\eta}r_t - \frac{w_t}{\eta}r_{t+1}\right] + C\eta + E\eta^2 + G\eta^4 \\
    &=    \left[\frac{r_0}{\eta W_T} - \frac{w_T}{\eta W_T}r_{T+1}\right] + C\eta + E\eta^2 + G\eta^4 \,.
\end{align*}
So we have: 
\begin{align*}
    \frac{1}{W_T}\sum_{t=0}^T w_t e_t + \frac{w_T}{\eta W_T}r_{T+1} \leq \frac{r_0}{\eta W_T} + C\eta + E\eta^2 + G\eta^4 \,.
\end{align*}
Also using the fact that $W_T \leq \frac{w_T}{a\eta}$ and $W_T \geq (1-A\eta)^{-(T+1)}$ we have: 
\begin{align*}
    \frac{1}{W_T}\sum_{t=0}^T w_t e_t + a r_{T+1} &\leq \frac{(1-A\eta)^{-(T+1)} r_0}{\eta}+ C\eta + E\eta^2 + G\eta^4 \\
    &\leq \frac{r_0 \exp{\left(-a\eta (T+1)\right)}}{\eta}+ C\eta + E\eta^2 + G\eta^4 \,.
\end{align*}
Now we consider two cases: 
\begin{itemize}
    \item[1)] $\frac 1 D \geq \frac{\ln\left(\max\left\{2,\frac{A^2T^2r_0}{C}\right\}\right)}{AT}$: We choose $\eta = \frac{\ln\left(\max\left\{2,\frac{A^2T^2r_0}{C}\right\}\right)}{AT}$. With this choice we get a rate of: 
\end{itemize}
\begin{align*}
    \frac{1}{W_T}\sum_{t=0}^T w_t e_t + A r_{T+1} &\leq \frac{r_0  A T}{\ln\left(\max\left\{2,\frac{A^2T^2r_0}{C}\right\}\right)} \exp\left(-\ln\left(\max\left\{2,\frac{A^2T^2r_0}{C}\right\}\right)\right) + \tilde{\mathcal{O}} \left(\frac{C}{AT} \right)\\
    &\quad + \tilde{\mathcal{O}} \left(\frac{E}{A^2T^2} \right) +  \tilde{\mathcal{O}} \left(\frac{G}{A^4T^4} \right) \\
    &\leq  \tilde{\mathcal{O}} \left(\frac{r_0 A}{T^2} \right)+ \tilde{\mathcal{O}} \left(\frac{C}{AT} \right) + \tilde{\mathcal{O}} \left(\frac{E}{A^2T^2} \right) +  \tilde{\mathcal{O}} \left(\frac{G}{A^4T^4} \right) \,.
\end{align*}
\begin{itemize}
    \item[2)] $\frac 1 D \leq \frac{\ln\left(\max\left\{2,\frac{A^2T^2r_0}{C}\right\}\right)}{AT}$: We choose $\eta = \frac 1 D$. With this choice we get a rate of: 
\end{itemize}
\begin{align*}
    \frac{1}{W_T}\sum_{t=0}^T w_t e_t + A r_{T+1} &\leq D r_0 \exp{\left(\frac{-A T}{D} \right)} + \frac{C}{D} + \frac{E}{D^2} + \frac{G}{D^4} \\
    &\leq D r_0 \exp{\left(\frac{- A T}{D} \right)}+ \tilde{\mathcal{O}} \left(\frac{C }{A T} \right) + \tilde{\mathcal{O}} \left(\frac{E}{A^2T^2} \right) +  \tilde{\mathcal{O}} \left(\frac{G}{A^4T^4} \right) \,.
\end{align*}   
\end{proof}

\subsection{Proof of Theorem \ref{thm: LSGD_upper_bound}} \label{app:strongly convex upper bound proof}
\begin{proof}
The proof follows a similar approach to \cite{yuan2020federated} while we assume that we have heterogeneity. In the following proof, we use the same notation as the work \cite{woodworth2020minibatch}. We drop the subscript $r$ for simplicity and use $t$ instead which is in the range $[0,KR-1]$. We use the notation $x^m_t$ as the parameters of client $m$ at time step $t$ and $g^m_t$ as the stochastic gradient on client $m$ at time step $t$. Also because $\beta = 1$, the update rule $x_{r+1} = x_r + \frac{\beta}{M} \sum_m (x^m_{r,K} - x_r)$ reduces to $x_{r+1} =  \frac{1}{M} \sum_m x^m_{r,K}$. We also define $\bar x_t = \frac 1 M \sum_{m=1}^M x^m_{t}$ as the average of parameters over all clients at time step $t$. Starting with the distance from the optimal point and taking the conditional expectation on the previous iterate $x^m_t, \forall m \in [M]$ we have: 
\begin{align}
    &\mathbb{E}_t \sb{\norm{\bar x_{t+1} - x^{\star}}^2} \nonumber\\
    &= \mathbb{E}_t \sb{\norm{\bar x_{t} - x^{\star} - \frac{\eta_t}{M} \sum_{m=1} ^{M} \nabla F_m(x^m_t) + \frac{\eta_t}{M} \sum_{m=1} ^{M} \nabla F_m(x^m_t) - \frac{\eta_t}{M} \sum_{m=1} ^{M} g^m_t}^2} \nonumber\\
    &\overset{(\text{Lemma \ref{lemma: noise of gradient differences}})}{\leq}  \norm{\bar x_{t} - x^{\star} - \frac{\eta_t}{M} \sum_{m=1} ^{M} \nabla F_m(x^m_t)}^2 + \frac{\eta_t^2 \sigma^2}{M} \nonumber\\
    &= \norm{\bar x_{t} - x^{\star} - \eta_t \nabla F(\bar x_t) + \eta_t \nabla F(\bar x_t) - \frac{\eta_t}{M} \sum_{m=1} ^{M} \nabla F_m(x^m_t)}^2 + \frac{\eta_t^2 \sigma^2}{M} \nonumber\\
    &\overset{(\text{Lemma \ref{Lemma: generalized triangle inequality}})}{\leq} \left(1+\frac{\eta_t \mu}{2}\right)\norm{\bar x_{t} - x^{\star} - \eta_t \nabla F(\bar x_t)}^2 + \eta_t^2\left(1+\frac{2}{\eta_t \mu}\right) \norm{\nabla F(\bar x_t) - \frac 1 M \sum_{m=1} ^{M} \nabla F_m(x^m_t)}^2+ \frac{\eta_t^2 \sigma^2}{M}  \,. \label{eq: first decrease lemma}
\end{align}
For the first term in (\ref{eq: first decrease lemma}) we have: 
\begin{align*}
    \norm{\bar x_{t} - x^{\star} - \eta_t \nabla F(\bar x_t)}^2 = \norm{\bar x_{t} - x^{\star}}^2 + \eta_t^2 \norm{\nabla F(\bar x_t)}^2 -2 \eta_t \Bigr \langle \bar x_{t} - x^{\star}, \nabla F(\bar x_t)  \Bigr \rangle \,.
\end{align*}
For the second term in the above equation we have: 
\begin{align*}
    \eta_t^2 \norm{\nabla F(\bar x_t)}^2 &= \eta_t^2 \norm{\nabla F(\bar x_t) - \nabla F(x^{\star})}^2 \\
    &\overset{(\text{Lemma \ref{lemma: co-coercivity of gradients}})}{\leq} 2H\eta_t^2 \Bigr[  F(\bar x_t) - F(x^{\star}) \Bigr] \,.
\end{align*}
For the third term in the equality we have: 
\begin{align*}
    -2 \eta_t \Bigr \langle \bar x_{t} - x^{\star}, \nabla F(\bar x_t)  \Bigr \rangle \leq - 2\eta_t \Bigr[ F(\bar x_t) - F(x^{\star}) \Bigr] - \eta_t \mu  \norm{\bar x_t - x^{\star}}^2
\end{align*}
Now by putting everything together we have: 
\begin{align*}
    &\norm{\bar x_{t} - x^{\star} - \eta_t \nabla F(\bar x_t)}^2 \\ 
    &= \norm{\bar x_{t} - x^{\star}}^2 + \eta_t^2 \norm{\nabla F(\bar x_t)}^2 -2 \eta_t \Bigr \langle \bar x_{t} - x^{\star}, \nabla F(\bar x_t)  \Bigr \rangle  \\
    &\leq \norm{\bar x_{t} - x^{\star}}^2 + 2H\eta_t^2 \Bigr[  F(\bar x_t) - F(x^{\star}) \Bigr] - 2\eta_t \Bigr[ F(\bar x_t) - F(x^{\star}) \Bigr] - \eta_t \mu  \norm{\bar x_t - x^{\star}}^2 \,.
\end{align*}
With the choice of $\eta_t \leq \frac{1}{2H}$ we have:
\begin{align*}
    \norm{\bar x_{t} - x^{\star} - \eta_t \nabla F(\bar x_t)}^2 \leq (1-\eta_t \mu) \norm{\bar x_{t} - x^{\star}}^2 - \eta_t \Bigr[ F(\bar x_t) - F(x^{\star}) \Bigr] \,.
\end{align*}
Then we multiply both sides by the coefficient $(1+\frac{\eta_t \mu}{2})$ and we have: 
\begin{align*}
    \left(1+\frac{\eta_t \mu}{2}\right)\norm{\bar x_{t} - x^{\star} - \eta_t \nabla F(\bar x_t)}^2 &\leq \left(1+\frac{\eta_t \mu}{2}\right)(1-\eta_t \mu) \norm{\bar x_{t} - x^{\star}}^2 - \eta_t \left(1+\frac{\eta_t \mu}{2}\right) \Bigr[ F(\bar x_t) - F(x^{\star}) \Bigr] \\
    &\leq \left(1-\frac{\eta_t \mu}{2}\right)\norm{\bar x_{t} - x^{\star}}^2 - \eta_t \Bigr[ F(\bar x_t) - F(x^{\star}) \Bigr]
\end{align*}
For the second term in (\ref{eq: first decrease lemma}) we have: 
\begin{align*}
    &\eta_t^2\left(1+\frac{2}{\eta_t \mu}\right) \norm{\frac 1 M \sum_{m=1} ^{M} \nabla F_m(x^m_t) - \nabla F(\bar x_t)}^2 \\
    &\leq \frac{4\eta_t}{ \mu} \norm{\frac 1 M \sum_{m=1} ^{M} \nabla F_m(x^m_t) - \nabla F(\bar x_t)}^2 \\
    &= \frac{4\eta_t}{ \mu}  \norm{\frac 1 M \sum_{m=1} ^{M}\Bigr(\nabla F_m(x^m_t) - \nabla F(x^m_t) + \nabla F(\bar x_t) - \nabla F_m(\bar x_t) \Bigr) + \frac 1 M \sum_{m=1} ^{M} \nabla F(x^m_t)-\nabla F(\bar x_t)}^2 \\
    & \overset{(\text{Lemma \ref{Lemma: jensen's inequality},\ref{Lemma: triangle inequality}})}{\leq}  \frac {8\eta_t}{\mu M} \sum_{m=1} ^{M} \norm{\nabla F_m(x^m_t) - \nabla F(x^m_t) + \nabla F(\bar x_t) - \nabla F_m(\bar x_t)}^2 + \frac{8\eta_t}{\mu} \norm{\frac 1 M \sum_{m=1} ^{M} \nabla F(x^m_t)-\nabla F(\bar x_t)}^2 \\
    &\overset{(\text{Lemma \ref{lemma: mime's inequality}})}{\leq} \frac{8\eta_t}{\mu} \tau^2 \varXi_t + \frac{8\eta_t}{\mu} \norm{\frac 1 M \sum_{m=1} ^{M} \nabla F(x^m_t)-\nabla F(\bar x_t)}^2  \,.
\end{align*}
For the second term in the above inequality we have: 
\begin{align*}
    & \frac{8\eta_t}{\mu} \norm{\frac 1 M \sum_{m=1} ^{M} \nabla F(x^m_t)-\nabla F(\bar x_t)}^2 \\
    &= \frac{8\eta_t}{\mu}  \norm{\frac 1 M \sum_{m=1} ^{M}\Bigr(\nabla F(x^m_t)-\nabla F(\bar x_t)-\nabla^2 F(\bar x_t)^{\top}(x^m_t-\bar x_t)\Bigr) + \underbrace{\frac 1 M \sum_{m=1} ^{M}\nabla^2 F(\bar x_t)^{\top}(x^m_t-\bar x_t)}_{=0}}^2 \\
    &= \frac{8\eta_t}{\mu}  \Biggr(\norm{\frac 1 M \sum_{m=1} ^{M}\Bigr(\nabla F(x^m_t)-\nabla F(\bar x_t)-\nabla^2 F(\bar x_t)^{\top}(x^m_t-\bar x_t)\Bigr)}\Biggr)^2 \\
    &\overset{(\text{Triangle Inequality})}{\leq} \frac{8\eta_t}{\mu}  \Biggr(\frac{1}{M}\sum_{m=1} ^{M}\norm{\nabla F(x^m_t)-\nabla F(\bar x_t)-\nabla^2 F(\bar x_t)^{\top}(x^m_t-\bar x_t)}\Biggr)^2 \\
    &\overset{(\text{Lemma \ref{lemma: quadratic function approximation inequality}})}{\leq} \frac{8\eta_t}{\mu}  \Biggr(\frac{1}{M}\sum_{m=1} ^{M} \frac Q 2 \norm{x^m_t-\bar x_t}^2 \Biggr)^2 \\
    &\overset{(\text{Jensen's Inequality})}{\leq} \frac{2Q^2 \eta_t}{\mu M} \sum_{m=1} ^{M} \norm{x^m_t-\bar x_t}^4
\end{align*}
Now by plugging everything back into (\ref{eq: first decrease lemma}) we have: 
\begin{align*}
    \mathbb{E}_t \sb{\norm{\bar x_{t+1} - x^{\star}}^2} \leq \left(1-\frac{\eta_t \mu}{2}\right)\norm{\bar x_{t} - x^{\star}}^2 - \eta_t \Bigr[ F(\bar x_t) - F(x^{\star}) \Bigr] + \frac{8 \tau^2\eta_t}{\mu}  \varXi_t + \frac{2Q^2 \eta_t}{\mu M} \sum_{m=1}^M \ee \sb{\norm{x^m_t - \bar x_t}^4}  + \frac{\eta_t^2 \sigma^2}{M} \,.
\end{align*}
Then we divide both sides by $\eta_t = \eta$, rearrange the terms and take the unconditional expectation, and we have: 
\begin{align*}
    &\mathbb{E} \Bigr[ F(\bar x_t) - F(x^{\star}) \Bigr] \\
    &\leq \left(\frac 1 \eta_t -\frac{ \mu}{2}\right) \mathbb{E} \sb{\norm{\bar x_{t} - x^{\star}}^2} - \frac 1 \eta_t \mathbb{E} \sb{\norm{\bar x_{t+1} - x^{\star}}^2} + \frac{8 \tau^2}{\mu}  \varXi_t + \frac{2Q^2 }{\mu M} \sum_{m=1}^M \ee \sb{\norm{x^m_t - \bar x_t}^4} + \frac{\eta \sigma^2}{M} \\
    &\overset{(\text{Lemma (\ref{lemma: weight variance upper bound},\ref{lem:cons_error_fourth})})}{\leq} \left(\frac 1 \eta -\frac{ \mu}{2}\right) \mathbb{E} \sb{\norm{\bar x_{t} - x^{\star}}^2} - \frac 1 \eta \mathbb{E} \sb{\norm{\bar x_{t+1} - x^{\star}}^2} + \frac{\eta \sigma^2}{M} + \\ 
    &\hspace{3.5cm}\frac{24 K \tau^2 \sigma^2 \eta^2}{\mu} + \frac{48 K^2 \tau^2   \zeta^2 \eta^2}{\mu} + \frac{11840 K^2 Q^2 \sigma^4 \eta^4}{\mu} + \frac{7680 Q^2 K^4   \zeta^4 \eta^4}{\mu}  \,.
\end{align*}
After tuning the stepsize using the Lemma \ref{lemma: stepsize} with the weights $w_t = (1-\frac{\mu \eta}{2})^{-(t+1)}$ and $W_T = \sum_{t=0}^{KR-1} w_t$ and by summing over $t=0,..,KR-1$ we have: 
\begin{align*}
     &\mathbb{E} \left[F\left(\frac{1}{W_T} \sum_{t=0}^{KR-1} w_t\bar x_t\right) - F(x^{\star})\right] + \frac{\mu}{2} \norm{\bar x_{T+1} - x^\star}^2 \\
     &\leq 2HB^2 \exp \left(-\frac{\mu T}{4H}\right) + \tilde{\mathcal{O}} \left( \frac{\sigma^2}{\mu M KR}\right)+ \tilde{\mathcal{O}} \left( \frac{\tau^2\sigma^2}{\mu^3 KR^2}\right)+ \tilde{\mathcal{O}} \left( \frac{\tau^2  \zeta^2}{\mu^3 R^2}\right)+ \tilde{\mathcal{O}} \left( \frac{Q^2\sigma^4}{\mu^5 K^2R^4}\right)+ \tilde{\mathcal{O}} \left( \frac{Q^2  \zeta^4}{\mu^5 R^4}\right) \,.
\end{align*}
It is worth mentioning that the notation $\tilde{\mathcal{O}} (.)$ means the bound holds up to some logarithmic factors appearing in the nominator for the last five terms. The logarithmic factor is in the form of $\ln\left(\max\left\{2,T^2\right\}\right)$ (some absolute constants are ignored). In the literature, it is also common to ignore these factors \citep{yuan2020federated, koloskova2020unified}.
\end{proof}


\subsection{Proof of Corollary \ref{col: convex upper bound}}
We use the regularization technique to derive the convergence rate for the convex case from the strongly-convex result. This technique is standard in the literature, see e.g.\ \citep{hazan2016introduction}. For the sake of completeness, we repeat the argument here.

Let $F(x)$ be a convex function. We construct a regularized version of this function $F_\mu(x)$ as:
\begin{align*}
    F_\mu(x) = F(x) + \frac{\mu}{2} \norm{x-x_0}^2    \,.
\end{align*}
Next we define: 
\begin{align*}
    x^\star_\mu &= \underset{x}{\arg\min} \hspace{1mm}F_\mu(x)  \,,\\
    x^\star &= \underset{x}{\arg\min}\hspace{1mm} F(x) \,.
\end{align*}
We have that $F_\mu(x^\star_\mu) \leq F_\mu(x^\star)$. Then we upper bound the function sub-optimality for the convex function $F(x)$:
\begin{align*}
    F(\bar x_t) - F(x^\star) &\leq F_\mu(\bar x_t) - \frac{\mu}{2} \norm{\bar x_t - x_0}^2 - F_\mu(x^\star) + \frac{\mu}{2} \norm{x^\star - x_0}^2\\
    &\leq F_\mu(\bar x_t)- F_\mu(x^\star)+ \frac{\mu}{2} \norm{x^\star - x_0}^2 \\
    &\leq F_\mu(\bar x_t)- F_\mu(x^\star_\mu)+ \frac{\mu}{2} \norm{x^\star - x_0}^2 \\
    &\leq F_\mu(\bar x_t)- F_\mu(x^\star_\mu)+ \frac{\mu}{2}B^2  \,.
\end{align*}
The last step is to tune the $\mu$. We assume that we want to achieve $\epsilon$-accuracy when running local SGD on the convex function $F$, so we have: 
\begin{align*}
    F(\bar x_t) - F(x^\star) &\leq \left(F_\mu(\bar x_t)- F_\mu(x^\star_\mu)\right)+ \frac{\mu}{2}B^2 \leq \epsilon \,.
\end{align*}
To satisfy this condition, it is enough to ensure that both terms in the rate are less than $\frac{\epsilon}{2}$. We solve this for the regularization term, and we get: 
\begin{align*}
    \frac{\mu}{2}B^2 \leq \frac{\epsilon}{2}\,,
\end{align*}
which results in:
\begin{align*}
    \mu \leq \frac{\epsilon}{B^2} \,.
\end{align*}
Now we proceed by replacing $\mu$ in all terms and deriving the rate.
\begin{proof}
By assuming $\mu=\frac{\epsilon}{B^2}$, we will find the condition on $K$ and $R$ to reach $\epsilon$-accuracy using Theorem \ref{thm: LSGD_upper_bound}. For simplicity, we drop all constant numbers which do not affect the rate. To recall, in the strongly convex case we had proven a rate of: 
\begin{align*}
    HB^2 \exp \left(-\frac{\mu KR}{H}\right) + \tilde{\mathcal{O}} \left( \frac{\sigma^2}{\mu M KR}\right)+ \tilde{\mathcal{O}} \left( \frac{\tau^2\sigma^2}{\mu^3 KR^2}\right)+ \tilde{\mathcal{O}} \left( \frac{\tau^2  \zeta^2}{\mu^3 R^2}\right)+ \tilde{\mathcal{O}} \left( \frac{Q^2\sigma^4}{\mu^5 K^2R^4}\right)+ \tilde{\mathcal{O}} \left( \frac{Q^2  \zeta^4}{\mu^5 R^4}\right) \,. 
\end{align*}
For the first term we solve the inequality: 
\begin{align*}
    HB^2 \exp \left(-\frac{\epsilon KR}{HB^2}\right) \leq \epsilon \quad \Longrightarrow \quad KR \geq \frac{HB^2}{\epsilon} \ln \left( \frac{\epsilon}{HB^2} \right) = \tilde{\mathcal{O}} \left( \frac{HB^2}{\epsilon}\right) \,.
\end{align*}
For the second term we solve the inequality:
\begin{align*}
    \frac{\sigma^2 B^2}{\epsilon M KR} \leq \epsilon \quad \Longrightarrow \quad KR \geq \frac{\sigma^2 B^2}{M\epsilon^2} \,.
\end{align*}
For the third term we solve the inequality: 
\begin{align*}
     \frac{\tau^2\sigma^2 B^6}{\epsilon^3 KR^2} \leq \epsilon \quad \Longrightarrow \quad KR^2 \geq \frac{\tau^2 \sigma^2 B^6}{\epsilon^4} \,.
\end{align*}
For the fourth term we solve the inequality:
\begin{align*}
    \frac{\tau^2   \zeta^2 B^6}{\epsilon^3  R^2} \leq \epsilon \quad \Longrightarrow \quad R^2 \geq \frac{\tau^2   \zeta^2 B^6}{\epsilon^4} \,.
\end{align*}
For the fifth term we solve the inequality:
\begin{align*}
    \frac{Q^2 \sigma^4}{\mu^5 K^2 R^4} \leq \epsilon \quad \Longrightarrow \quad K^2R^4 \geq \frac{Q^2 \sigma^4 B^{10}}{\epsilon^6} \,.
\end{align*}
And for the last term: 
\begin{align*}
    \frac{Q^2  \zeta^4 B^{10}}{\epsilon^5 R^4} \leq \epsilon \quad \Longrightarrow \quad R^4 \geq \frac{Q^2   \zeta^4 B^{10}}{\epsilon^6} \,.
\end{align*}
Finally, the rate would be the maximum of above terms which we can write: 
\begin{align*}
    &\tilde{\mathcal{O}} \left (\max\left\{\frac{HB^2}{\epsilon},\frac{\sigma^2 B^2}{M\epsilon^2}, \frac{\tau^2 \sigma^2 B^6}{\epsilon^4},\frac{\tau^2   \zeta^2 B^6}{\epsilon^4},\frac{Q^2 \sigma^4 B^{10}}{\epsilon^6},\frac{Q^2   \zeta^4 B^{10}}{\epsilon^6}\right\}\right) \leq \\
    &\tilde{\mathcal{O}} \left (\frac{HB^2}{\epsilon}+\frac{\sigma^2 B^2}{M\epsilon^2}+ \frac{\tau^2 \sigma^2 B^6}{\epsilon^4}+\frac{\tau^2   \zeta^2 B^6}{\epsilon^4}+\frac{Q^2 \sigma^4 B^{10}}{\epsilon^6}+\frac{Q^2   \zeta^4 B^{10}}{\epsilon^6} \right) \,.
\end{align*}
Or in terms of $K$ and $R$:
    \begin{align*}
        \tilde{\mathcal{O}} \left ( \frac{HB^2}{KR} + \frac{\sigma B}{\sqrt{MKR}} + \frac{(\tau \sigma B^3)^{1/2}}{K^{1/4}R^{1/2}} + \frac{(\tau  \zeta B^3)^{1/2}}{R^{1/2}}
      + \frac{(Q\sigma^2B^5)^{1/3}}{K^{1/3}R^{2/3}} + \frac{(Q  \zeta^2B^5)^{1/3}}{R^{2/3}} \right) \,.
    \end{align*}
It's worth mentioning that during the process of deriving rate for convex regime from strongly convex regime, we also ignore some logarithmic factors which is why we used the notation $\tilde{\mathcal{O}}(.)$. In fact the term $\frac{\ln\left(\max\left\{2,\frac{M \mu^2T^2B^2}{\sigma^2}\right\}\right)}{\mu T}$ (some constants are dropped for simplicity) that we use for tuning the stepsize depends on $\mu$. By the choice of $\mu = \frac{\epsilon}{B^2}$ this term becomes $\frac{B^2\ln\left(\max\left\{2,\frac{M \epsilon^2T^2}{B^2 \sigma^2}\right\}\right)}{\epsilon T}$ which will be used to upper bound the terms in our rate which are in the form of $\frac{C}{D} + \frac{E}{D^2} + \frac{G}{D^4}$ (refer to \ref{app:lemmas}). We use the fact that $\frac 1 D \leq \frac{B^2\ln\left(\max\left\{2,\frac{M \epsilon^2T^2}{B^2 \sigma^2}\right\}\right)}{\epsilon T}$. If $2 \geq \frac{M \epsilon^2T^2}{B^2 \sigma^2}$, then there is no extra logarithmic factor appearing in our rate. Only in the case of $2 \leq \frac{M \epsilon^2T^2}{B^2 \sigma^2}$ we get an extra factor of $\ln(T^2)$. In this case, for the term $\frac{C}{D}$ as an example we have: 
\begin{align*}
    \frac{C}{D} \leq \frac{CB^2\ln\left(\frac{M \epsilon^2T^2}{B^2 \sigma^2}\right)}{\epsilon T}
\end{align*}
And to achieve $\epsilon$-accuracy we want this term to be less that $\epsilon$ so we have: 
\begin{align*}
    \frac{CB^2\ln\left(\frac{M \epsilon^2T^2}{B^2 \sigma^2}\right)}{\epsilon T} \leq \epsilon
\end{align*}
So we need to set $T$ in the following way:
\begin{align*}
    T &\geq \frac{CB^2\ln\left(\frac{M \epsilon^2T^2}{B^2 \sigma^2}\right)}{\epsilon^2} \\
    &\geq \frac{CB^2\ln\left(\frac{M \epsilon^2}{B^2 \sigma^2}\right)}{\epsilon^2} = \tilde{\mathcal{O}}(\frac{1}{\epsilon^2})
\end{align*}
And for the other terms we do the same. 

\end{proof}


\section{Missing Details from Section \ref{sec:fixed}}\label{app:fixed}
\subsection{Proof of Proposition \ref{prop:lsgd_fixed_point}}
Assuming the local SGD algorithm converges, i.e., the hyper-parameters are set to achieve that and $R\to \infty$, we would like to calculate its fixed point. Let's denote the fixed point by $x_\infty$. Then $x_\infty$ must satisfy the following requirements,
\begin{align*}
    x_\infty = x_\infty + \frac{\beta}{M}\sum_{m\in[M]}\Delta^m(x_\infty) \equiv \sum_{m\in[M]}\Delta^m(x_\infty) =0,
\end{align*}
where $\Delta^m(x_\infty)$ is the update on machine $m$ for a communication round starting at the fixed point $x_\infty$. Note that the above equation does not depend on $\beta$. Also, note that this is very similar to the average drift assumption of \citet{wang2022unreasonable}. Unwinding the update, we get the following,
\begin{align*}
    &\sum_{m\in[M]}\sum_{k\in[K]}A_m(I-\eta A_m)^{k-1}(x_\infty - x_m^\star) =0,\\
    &\Leftrightarrow^{\text{if $\eta=0$}} x_\infty = \frac{1}{M}\sum_{m\in[M]}A^{-1}A_mx_m^\star = x^\star,\\
    &\Leftrightarrow^{\text{if $\eta>0$}} x_\infty = \frac{1}{M}\sum_{m\in[M]}C^{-1}C_mx_m^\star,
\end{align*}
where $C_m := I- (I-\eta C_m)^K$, and $C := \frac{1}{M}\sum_{m\in[M]}C_m$. Note that $x_\infty(\eta, K)$ is a function of $\eta, K$ and is unaffected by the choice of $\beta$. The above proof also highlights a key limitation of the assumption \ref{ass:rho}; they assume that local SGD is going to converge to $x^\star$ as opposed to any other fixed point, defining the average drift at the optimum. This makes their analysis more complicated than it should have been. 

\subsection{Proof of Proposition \ref{prop:bar_star_distance}}
\begin{proof}
    We first bound the distance between $x^\star$ and $\bar x^\star$,
    \begin{align*}
        \norm{x^\star - \bar x^\star} &= \norm{\frac{1}{M}\sum_{m\in[M]}\rb{I-A^{-1}A_m}x_m^\star},\\
        &= \norm{\frac{1}{M}\sum_{m\in[M]}A^{-1}\rb{A-A_m}(x_m^\star-x^\star)},\\
        &\leq \frac{1}{M}\sum_{m\in[M]} \norm{A^{-1}\rb{A-A_m}(x_m^\star-x^\star)},\\
        &\leq \frac{1}{M}\sum_{m\in[M]} \norm{A^{-1}}\norm{A-A_m}\norm{x_m^\star-x^\star},\\
        &\leq \frac{1}{M}\sum_{m\in[M]} \frac{1}{\mu}\cdot\tau \cdot \norm{x_m^\star-x^\star},\\
        &= \frac{\tau \zeta_\star}{H\mu}.
    \end{align*}
    We now bound the distance between $x_\infty(K>1, \eta)$ and $\bar x^\star$ similarly,
    \begin{align*}
        \norm{x_\infty - \bar x^\star} &= \norm{\frac{1}{M}\sum_{m\in[M]}\rb{I-C^{-1}C_m}x_m^\star},\\
        &= \norm{\frac{1}{M}\sum_{m\in[M]}C^{-1}\rb{C-C_m}(x_m^\star-x^\star)},\\
        &\leq \frac{1}{M}\sum_{m\in[M]} \norm{C^{-1}\rb{C-C_m}(x_m^\star-x^\star)},\\
        &\leq \frac{1}{M}\sum_{m\in[M]} \norm{C^{-1}}\norm{C-C_m}\norm{x_m^\star-x^\star},\\
        &\leq \frac{\zeta_\star}{H}\norm{C^{-1}}\frac{1}{M}\sum_{m\in[M]} \norm{C-C_m},\\
        &= \frac{\zeta_\star}{H}\cdot\frac{1}{\lambda_{min}(C)}\frac{1}{M}\sum_{m\in[M]} \norm{C-C_m},\\
        &\leq \frac{\zeta_\star}{H}\cdot\frac{1}{1-(1-\eta \inf_{m\in[M]}\lambda_{min}(A_m))^K}\frac{1}{M}\sum_{m\in[M]} \norm{C-C_m},\\
        &\leq \frac{\zeta_\star}{H}\cdot\frac{1}{1-(1-\eta \mu)^K}\frac{1}{M^2}\sum_{m,n\in[M]} \norm{C_n-C_m},\\
        &\leq \frac{\zeta_\star}{H}\cdot\frac{1}{1-(1-\eta \mu)^K}\sup_{m,n\in[M]} \norm{(I-\eta A_m)^K-(I-\eta A_n)^K},\\
        &\leq \frac{\zeta_\star}{H}\cdot\frac{1}{1-(1-\eta \mu)^K}\sup_{m,n\in[M]} \norm{\eta(A_m-A_n)}K(1-\eta\mu)^{K-1},\\
        &\leq \frac{\zeta_\star}{H}\cdot\frac{\tau \eta K(1-\eta\mu)^{K-1}}{1-(1-\eta \mu)^K},
    \end{align*}
    where to get the second last inequality, we note the following sequence of inequalities with a clever adding and subtracting of several terms,
    \begin{align*}
        (I-\eta A_m)^K-(I-\eta A_n)^K &= (I-\eta A_m)^K - \sum_{k\in[K]}(I-\eta A_n)^{k}(I-\eta A_m)^{K-k}\\
        &\quad+ \sum_{k\in[K]}(I-\eta A_n)^{k}(I-\eta A_m)^{K-k} - (I-\eta A_n)^K\enspace,\\
        &= (I-\eta A_m)^K  - \sum_{k\in[K]}(I-\eta A_n)^{k}(I-\eta A_m)^{K-k} + \sum_{k\in[K-1]}(I-\eta A_n)^{k}(I-\eta A_m)^{K-k}\\
        &\quad + (I-\eta A_n)^K - (I-\eta A_n)^K\enspace,\\
        &= (I-\eta A_m)^K + \sum_{k\in[K-1]}(I-\eta A_n)^{k}(I-\eta A_m)^{K-k} - \sum_{k\in[K]}(I-\eta A_n)^{k}(I-\eta A_m)^{K-k}\enspace,\\
        &= \sum_{k\in[0, K-1]}(I-\eta A_n)^{k}(I-\eta A_m)^{K-k} - \sum_{k\in[K]}(I-\eta A_n)^{k}(I-\eta A_m)^{K-k}\enspace,\\
        &= \sum_{k\in[K]}(I-\eta A_n)^{k-1}(I-\eta A_m)^{K-k+1} - \sum_{k\in[K]}(I-\eta A_n)^{k}(I-\eta A_m)^{K-k}\enspace,\\
        &= \sum_{k\in[K]}\sb{(I-\eta A_n)^{k-1}(I-\eta A_m)^{K-k+1} - (I-\eta A_n)^{k}(I-\eta A_m)^{K-k}}\enspace,\\
        &= \sum_{k\in[K]}\sb{\eta(I-\eta A_n)^{k-1}(A_n-A_m)(I-\eta A_m)^{K-k}},
    \end{align*}
    which upon taking the norm implies, 
    \begin{align*}
        \norm{ (I-\eta A_m)^K-(I-\eta A_n)^K} &\leq \sum_{k\in[K]}\eta \norm{I-\eta A_n}^{k-1}\norm{A-n-A_m}\norm{I-\eta A_m}^{K-k},\\
        &\leq \eta \sum_{k\in[K]}(1-\eta\mu)^{k-1}\tau (1-\eta\mu)^{K-k},\\
        &\leq \eta \tau K(1-\eta\mu)^{K-1},
    \end{align*}
    which finishes the proof of the inequality and hence the proof as well.
\end{proof}

\subsection{Proof of Theorem \ref{thm:fixed_convergence}}

Having identified the convergence point of local SGD in the previous section as $x_\infty(\eta, K)$, in this section, we will characterize the rate of convergence to this point. 

\begin{proof}
    We first write the inner loop iterated on machine $m$ as follows,
    \begin{align*}
        x_{r,K}^m &= x_{r,K-1}^m - \eta\rb{A_m(x_{r, K-1}^m-x_m^\star)},\\
        &= x_m^\star + \rb{I-\eta A_m}^K\rb{x_{r,0}^m-x_m^\star},\\
        &= x_m^\star + \rb{I-\eta A_m}^K\rb{x_{r-1}-x_m^\star}.
    \end{align*}
    Subtracting $x_{r-1}$ from both sides,
    \begin{align*}
        x_{r,K}^m - x_{r-1} &= x_m^\star - x_{r-1} + \rb{I-\eta A_m}^K\rb{x_{r-1}-x_m^\star},\\
        &= \rb{I-\rb{I-\eta A_m}^K}\rb{x_m^\star - x_{r-1}}. 
    \end{align*}
With this in hand, we can write the update across communication rounds as follows,
    \begin{align*}
        x_r &= x_{r-1} + \frac{\beta}{M}\sum_{m\in[M]}\rb{x_{r,K}^m - x_{r-1}},\\
        &= x_{r-1} + \frac{\beta}{M}\sum_{m\in[M]}\rb{I-\rb{I-\eta A_m}^K}\rb{x_m^\star - x_{r-1}}.
    \end{align*}
    Subtracting the fixed point $\hat x$ on both sides, we get,
    \begin{align*}
        x_r -\hat x &= x_{r-1} -\hat x + \frac{\beta}{M}\sum_{m\in[M]}\rb{I-\rb{I-\eta A_m}^K}\rb{x_m^\star - x_{r-1}},\\
        &= x_{r-1} -\hat x + \frac{\beta}{M}\sum_{m\in[M]}C_m\rb{x_m^\star - x_{r-1}},\\
        &= \rb{I-\beta C}x_{r-1} - x_{\infty} + \frac{\beta B}{M}\sum_{m\in[M]}C^{-1}C_mx_m^\star,\\
        &= \rb{I-\beta C}x_{r-1} -  x_\infty + \beta C x_\infty,\\
        &= \rb{I-\beta C}\rb{x_{r-1}- x_\infty}.
    \end{align*}
    Unrolling this recursion for $R$ communication rounds,
    \begin{align*}
        x_R - x_\infty = \rb{I-\beta B}^R\rb{x_0-x_{\infty}}.
    \end{align*}
    Noting the initialization $x_0=0$ we get,
    \begin{align*}
        x_R &= \rb{I-\rb{I-\beta C}^R}x_\infty,\\
            &= \rb{I-\rb{I-\frac{\beta}{M}\sum_{m\in[M]}\rb{I-\rb{I-\eta A_m}^K}}^R}x_\infty,\\
            &= \rb{I-\rb{(1-\beta)\cdot I + \frac{\beta}{M}\sum_{m\in[M]}\rb{I-\eta A_m}^K}^R}x_\infty.
    \end{align*}
    In particular, we can bound the distance between $x_R$ and $x_\infty$ as follows,
    \begin{align*}
        \norm{x_R-x_\infty} &= \norm{\rb{I-\beta C}^R\rb{x_0-x_\infty}},\\
        &\leq \norm{I-\beta C}^R\norm{x_\infty},\\
        &=  \norm{I-\beta \cdot \frac{1}{M}\sum_{m\in[M]}\rb{I-\rb{I-\eta A_m}^K}}^R\norm{x_\infty},
    \end{align*}
    If we ensure that $\beta < 1/\norm{C}$, then the distance contracts with more communication rounds. Note that,
    \begin{align*}
        \norm{C} &= \norm{\frac{1}{M}\sum_{m\in[M]}\rb{I-\rb{I-\eta A_m}^K}},\\
        &= \norm{I - \frac{1}{M}\sum_{m\in[M]}\rb{I-\eta A_m}^K},\\
        &\leq \max_{m\in[M]}\norm{I-(I-\eta A_m)^K},\\
        &\leq 1-(1-\eta H)^K.
    \end{align*}
    This implies that we will have contraction with communication if $\beta < \frac{1}{1-(1-\eta H)^K}$. In this regime, we can calculate the overall convergence as follows,
    \begin{align*}
         \norm{x_R-x_\infty} &\leq \norm{I-\beta C}^R\norm{x_\infty},\\
         &\leq (1-\beta \norm{C^{-1}}^{-1})^R\norm{x_\infty},\\
         &\leq \rb{1-\beta\rb{1-\rb{1-\eta \mu}^K}}^R\norm{x_\infty},\\
         &\leq \rb{1-\beta\rb{1-\rb{1-\eta \mu}^K}}^R\norm{C^{-1}}\norm{\frac{1}{M}\sum_{m\in[M]}C_mx_m^\star},\\
         &\leq \rb{1-\beta\rb{1-\rb{1-\eta \mu}^K}}^R\cdot \frac{1-\rb{1-\eta L}^K}{1-\rb{1-\eta \mu}^K}B\enspace.  
    \end{align*}
    We can note that when $\beta=1$ this simplifies to,
    \begin{align*}
         \norm{x_R-x_\infty} &\leq \rb{1-\eta\mu}^{KR}\cdot\frac{1-\rb{1-\eta L}^K}{1-\rb{1-\eta \mu}^K}B\enspace.
    \end{align*}
    When $\beta = \frac{1}{c}\cdot \frac{1}{1-(1-\eta L)^K}$ for some $c> 1$ this reduces to,
    \begin{align*}
        \norm{x_R-x_\infty} &\leq \rb{1-\frac{1}{c}\cdot\frac{1-\rb{1-\eta \mu}^K}{1-\rb{1-\eta L}^K}}^R\cdot \frac{1-\rb{1-\eta L}^K}{1-\rb{1-\eta \mu}^K}B,\\
        &= \rb{1-\frac{1}{c\kappa'}}^R\kappa'B,
    \end{align*}
    where we define, 
    $$\kappa' := \frac{1-\rb{1-\eta L}^K}{1-\rb{1-\eta \mu}^K}.$$
    This finishes the convergence proof for local GD. To finish the proof of the theorem, we combine this result with the fixed point discrepancy implied by Proposition \ref{prop:bar_star_distance}. 
\end{proof}

\end{document}